\documentclass[conference,onecolumn]{IEEEtran}
\usepackage{amsmath,amssymb,amsthm,booktabs,graphicx,cite,tikz}
\usetikzlibrary{positioning,arrows.meta,fit}
\usepackage[section]{placeins}
\usepackage{url}

\IfFileExists{tables/numbers.tex}{% Generated by scripts/zz_numbers.py from the S5 scripts. Do not edit by hand.
\newcommand{\exactLagNvda}{86}
\newcommand{\trueSelloffNvda}{-47.1}
\newcommand{\ledgerSelloffNvda}{-39.0}

\newcommand{\trueSelloffNdx}{-19.7}
\newcommand{\ledgerSelloffNdx}{-15.1}

\newcommand{\alphaGTTwoNext}{+5.62}
\newcommand{\alphaLoGTTwoNext}{+2.31}
\newcommand{\alphaHiGTTwoNext}{+6.55}

\newcommand{\betaMeTwoNext}{0.84}
\newcommand{\alphaMeTwoNext}{+4.51}
\newcommand{\alphaLoMeTwoNext}{-0.17}
\newcommand{\alphaHiMeTwoNext}{+6.10}
\newcommand{\factorEnd}{2026-09-30}

\newcommand{\ffAlphaMe}{+3.96}
\newcommand{\ffAlphaLoMe}{+1.35}
\newcommand{\ffAlphaHiMe}{+6.57}

\newcommand{\ffAlphaGTTwo}{+2.62}
\newcommand{\ffAlphaLoGTTwo}{-1.70}
\newcommand{\ffAlphaHiGTTwo}{+6.94}

\newcommand{\ffAlphaMeTwo}{-0.67}
\newcommand{\ffAlphaLoMeTwo}{-4.44}
\newcommand{\ffAlphaHiMeTwo}{+3.09}

\newcommand{\pfPnlGTTwo}{-2.37}
\newcommand{\pfCostGTTwo}{57.5}
\newcommand{\pfAlphaAllGTTwo}{+51}
\newcommand{\pfTAllGTTwo}{+2.22}

\newcommand{\pfPnlMeTwo}{+4.99}
\newcommand{\pfCostMeTwo}{36.0}
\newcommand{\pfAlphaAllMeTwo}{+51}
\newcommand{\pfTAllMeTwo}{+2.00}
\newcommand{\pfBetaAllMeTwo}{1.63}
\newcommand{\pfAlphaLoMeTwo}{+1}
\newcommand{\pfAlphaHiMeTwo}{+100}

\newcommand{\pfAlphaAllMeTwoExMuAmd}{-8}
\newcommand{\pfTAllMeTwoExMuAmd}{-0.48}

\newcommand{\meTwoPnl}{+4.99}
\newcommand{\meTwoCost}{36.0}
\newcommand{\meTwoPnlPct}{+13.9}
\newcommand{\meTwoPnlMU}{+2.74}
\newcommand{\meTwoPnlAMD}{+2.50}
\newcommand{\meTwoPnlRest}{-0.24}
\newcommand{\pfLag}{5}
\newcommand{\betaMeTwoBuys}{+1.66}
\newcommand{\nMeTwoBuys}{72}
\newcommand{\betaMeTwoSells}{-1.56}
\newcommand{\nMeTwoSells}{53}
\newcommand{\pfDaysMeTwo}{184}
\newcommand{\pfDaysGTTwo}{184}
\newcommand{\pfDaysAll}{194}
\newcommand{\plNLots}{72}
\newcommand{\plDraws}{2000}

\newcommand{\plNTickers}{17}

\newcommand{\plTickAlphaMed}{+12}
\newcommand{\plTickAlphaHi}{+40}
\newcommand{\plTickPAlpha}{0.010}

\newcommand{\plTickPPnl}{0.009}

\newcommand{\attMeTwoMUPnl}{+2.74}

\newcommand{\attMeTwoAMDPnl}{+2.50}
\newcommand{\attMeTwoNVDALots}{5}

\newcommand{\attMeTwoNVDAPnl}{-0.22}
\newcommand{\attMeTwoGOOGLots}{8}

\newcommand{\attMeTwoGOOGPnl}{-0.30}
\newcommand{\attMeTwoTSMLots}{5}

\newcommand{\attMeTwoTSMPnl}{+0.02}

\newcommand{\attGTTwoMUPnl}{+2.11}

\newcommand{\attGTTwoAMDPnl}{+1.97}
\newcommand{\attGTTwoNVDALots}{19}

\newcommand{\attGTTwoNVDAPnl}{-2.02}
\newcommand{\attGTTwoGOOGLots}{15}

\newcommand{\attGTTwoGOOGPnl}{-1.58}
\newcommand{\attGTTwoTSMLots}{14}

\newcommand{\attGTTwoTSMPnl}{-0.90}
\newcommand{\prodLotK}{500}
\newcommand{\prodCap}{5}
\newcommand{\prodPnlClaudeRaw}{-63.86}
\newcommand{\prodPnlHClaudeRaw}{-97.93}
\newcommand{\prodCostClaudeRaw}{544.2}

\newcommand{\prodAlphaAllClaudeRaw}{+14}

\newcommand{\prodPnlClaudeCapExit}{+1.47}
\newcommand{\prodPnlHClaudeCapExit}{-7.96}
\newcommand{\prodCostClaudeCapExit}{114.4}

\newcommand{\prodAlphaAllClaudeCapExit}{+8}

\newcommand{\prodAlphaHGTTwoRaw}{-16}

\newcommand{\prodAlphaHGTTwoCap}{-11}

\newcommand{\prodAlphaAllGTTwoCapExit}{+54}

\newcommand{\prodAlphaHGTTwoCapExit}{-25}

\newcommand{\prodNVariants}{15}
\newcommand{\prodNSrc}{5}

\newcommand{\prodCapExitWinsH}{5}
\newcommand{\prodCapExitRetWins}{5}
\newcommand{\prodNLossH}{15}
\newcommand{\prodNNegAlphaH}{14}
\newcommand{\prodCapBeatsExitH}{4}
\newcommand{\plHNLots}{22}
\newcommand{\plHAlphaReal}{+20}
\newcommand{\plHPnlReal}{-0.44}
\newcommand{\plHAlphaMed}{-1}

\newcommand{\plHPAlpha}{0.28}

\newcommand{\ssTDeepSeekAll}{-2.11}
\newcommand{\ssCorrDeepSeekAll}{-0.35}

\newcommand{\ssTCtlDeepSeekAll}{-2.26}
\newcommand{\ssMedCallsDeepSeekAll}{4}

\newcommand{\ssTDeepSeekH}{-1.50}

\newcommand{\ssTCtlDeepSeekH}{-2.77}

\newcommand{\ssTClaudeH}{+3.44}

\newcommand{\ssTCtlClaudeH}{+0.77}

\newcommand{\nCallsClaude}{1109}
\newcommand{\nSellClaude}{20}
\newcommand{\pctSellClaude}{1.8}
\newcommand{\nCallsGemini}{685}
\newcommand{\nSellGemini}{58}
\newcommand{\pctSellGemini}{8.5}
\newcommand{\nCallsDeepSeek}{746}
\newcommand{\nSellDeepSeek}{205}
\newcommand{\pctSellDeepSeek}{27.5}

\newcommand{\nCallsGTTwo}{115}

\newcommand{\nCallsMeTwo}{145}

\newcommand{\pctSellMeTwo}{50.3}
\newcommand{\nThreeWay}{189}

\newcommand{\pctAllBuyCalm}{97.0}
\newcommand{\pctAllSellCalm}{1.0}

\newcommand{\pctAllBuyFall}{95.6}
\newcommand{\pctAllSellFall}{2.9}

\newcommand{\nTickers}{24}
\newcommand{\nTickersStart}{15}
\newcommand{\sampleStart}{2026-02-17}
\newcommand{\sampleEnd}{2026-11-24}
\newcommand{\nNdxCalls}{23}
\newcommand{\rhoErrCalm}{0.97}

\newcommand{\accMajCalm}{62.0}
\newcommand{\accJuryIndepCalm}{67.7}
\newcommand{\kappaMeanCalm}{0.66}
\newcommand{\rhoErrFall}{0.98}

\newcommand{\accMajFall}{65.6}
\newcommand{\accJuryIndepFall}{73.3}
\newcommand{\kappaMeanFall}{0.86}
\newcommand{\rhoErrAll}{0.96}
\newcommand{\nEffAll}{1.03}
\newcommand{\accMajAll}{59.9}
\newcommand{\accJuryIndepAll}{64.9}
\newcommand{\kappaMeanAll}{0.64}
\newcommand{\nAllBuyClsCalm}{97}

\newcommand{\accAllBuyCalm}{61.9}
\newcommand{\gapAllBuyCalm}{+8.6}

\newcommand{\baseUpCalm}{53.2}
\newcommand{\nAllBuyClsFall}{58}

\newcommand{\accAllBuyFall}{69.0}
\newcommand{\gapAllBuyFall}{+7.0}

\newcommand{\baseUpFall}{61.9}

\newcommand{\accClaudeCalm}{48.1}
\newcommand{\accClaudeFall}{62.8}
\newcommand{\accGeminiCalm}{49.7}
\newcommand{\accGeminiFall}{64.0}

\newcommand{\accMeTwoBuyCalm}{58.5}

\newcommand{\accMeTwoSellCalm}{63.3}

\newcommand{\accMeTwoBuyFall}{92.3}
\newcommand{\nMeTwoBuyFall}{26}
\newcommand{\accMeTwoSellFall}{52.2}
\newcommand{\nMeTwoSellFall}{23}
\newcommand{\betaClaude}{1.61}

\newcommand{\betaGemini}{1.50}

\newcommand{\betaDeepSeek}{1.20}

\newcommand{\betaMe}{1.54}

\newcommand{\betaGTTwo}{1.99}
\newcommand{\alphaGTTwo}{+4.99}
\newcommand{\alphaLoGTTwo}{+0.92}
\newcommand{\alphaHiGTTwo}{+6.13}
\newcommand{\betaMeTwo}{0.69}
\newcommand{\alphaMeTwo}{+4.90}
\newcommand{\alphaLoMeTwo}{-0.84}
\newcommand{\alphaHiMeTwo}{+6.13}
\newcommand{\meTwoEntryTotal}{+8.89}
\newcommand{\meTwoEntryMarket}{+3.59}
\newcommand{\meTwoEntrySelection}{+5.30}
\newcommand{\meTwoExitTotal}{+1.51}
\newcommand{\meTwoExitMarket}{+0.02}
\newcommand{\meTwoExitSelection}{+1.49}
\newcommand{\permObs}{58.5}
\newcommand{\permNull}{51.4}
\newcommand{\permP}{0.16}
\newcommand{\nExits}{53}
\newcommand{\cfNDs}{86}
\newcommand{\cfMeanDs}{+14.93}
\newcommand{\cfBetaDs}{3.21}
\newcommand{\cfAlphaDs}{+5.50}
\newcommand{\cfAlphaLoDs}{+1.02}
\newcommand{\cfAlphaHiDs}{+7.16}
\newcommand{\cfHoldDs}{38.2}
\newcommand{\cfPerRowDs}{+0.39}

\newcommand{\cfLedgerMatch}{72}
\newcommand{\cfLedgerBuys}{72}
\newcommand{\cfReplayBuys}{86}

\newcommand{\cfMeanCl}{+3.56}

\newcommand{\cfMeanGe}{+8.94}

\newcommand{\cfMeanAny}{+13.28}

\newcommand{\cfPerRowAny}{+0.37}

\newcommand{\cfMeanNone}{+4.18}
\newcommand{\cfBetaNone}{2.94}

\newcommand{\cfHoldNone}{115.5}
\newcommand{\cfPerRowNone}{+0.04}

\newcommand{\cfMeanAgeonly}{+7.91}
\newcommand{\cfBetaAgeonly}{2.00}

\newcommand{\cfPerRowAgeonly}{+0.40}

\newcommand{\grossGTTwoAll}{+8.06}
\newcommand{\netGTTwoAll}{+7.86}

\newcommand{\grossMeTwoAll}{+5.76}
\newcommand{\netMeTwoAll}{+5.56}
\newcommand{\costBp}{20}

\newcommand{\dsrSharpe}{0.42}
\newcommand{\dsrLots}{86}
\newcommand{\dsrNMin}{8}

\newcommand{\dsrMin}{1.000}

\newcommand{\dsrTwenty}{0.994}

\newcommand{\dsrFifty}{0.951}
}{}
\newcommand{\meTwo}{Me2}
\newcommand{\gtTwo}{GT-2}
\title{Unanimously Bullish: Correlated Bias Among Independent LLM Trading Signals, and Why Per-Call and
Portfolio Scoring Disagree}
\author{\IEEEauthorblockN{Zhang Qiyu}\IEEEauthorblockA{School of Information Technology, Nanyang Polytechnic, Singapore}}
\begin{document}
\maketitle
\begin{abstract}
Trading systems that combine several large language models assume the models err independently. We test this on
a live ledger of daily calls from Claude, Gemini and DeepSeek on large US-listed stocks. When all three call the same
stock, \pctAllBuyCalm\,\% of calls in calm markets and \pctAllBuyFall\,\% after a fall are unanimous Buys. Their
errors are highly correlated ($\rho=\rhoErrAll$), so they act as $\nEffAll$ effective voters, and their majority
is right \accMajAll\,\% of the time against \accJuryIndepAll\,\% for independent voters. A correlated-jury
proposition gives this loss in closed form. The dominant all-Buy outcome is right a few points more often than the
market rises, and a simple model shows that a strong shared bias drives that gap to zero. We then ask where
the edge of a role-split policy comes from; the policy buys on a game-theoretic combination of the models and
exits on the one model that dissents. The answer depends on how it is scored. Per call, it has the lowest market
beta (\betaMeTwo, because its Sells enter with negative beta), its exits cannot be told apart from random exits on the same dates ($p=\permP$), and whether its alpha
excludes zero depends on the interval method. As the dollar portfolio its trades imply, its invested dollars have a beta of
\pfBetaAllMeTwo{} and an alpha against the Nasdaq-100 of \pfAlphaAllMeTwo\,\% a year ($t=\pfTAllMeTwo$, not
deflated), equal to that of its entry source; no difference in alpha per dollar from the exits is detectable, though they limit the capital at risk, and
most of the edge comes from two stocks.
All results are in sample, since the rules were developed on the same data. Per-call and portfolio scoring can thus give different
verdicts on the same policy. For production, position limits help every signal source as risk control, and
DeepSeek's share of Sells is a borderline signal of the market's next move.
\end{abstract}
\section{Introduction}\label{sec:intro}

Several trading systems now ask more than one large language model (LLM) for a Buy or Sell call and combine the
answers. Why? The usual argument is the jury argument: if each model is right more often than not and the models
err independently, a majority vote is more accurate than any one of them. Recent work warns that this argument
needs independent voters, and that models that share training data tend to fail on the same questions
\cite{morante2026correlation}; in medical question answering, unanimous consensus can still be wrong
\cite{wang2026unanimously}. We measure how far that warning applies to live trading signals, and what it implies
for a policy that splits roles between models.

We study a ledger of daily calls on \nTickers{} large US-listed stocks from three LLMs (Claude, Gemini and
DeepSeek), from the author's own manual calls (Me), from a game-theoretic combination of the LLMs (\gtTwo) and from
a role-split policy (\meTwo) that enters on a \gtTwo{} Buy and exits on a DeepSeek Sell. The \meTwo{} rules were
developed on the same ledger, so its results are in sample.

The three LLMs are bullish together. When all three call the same stock on the same date, \pctAllBuyCalm\,\% of
the calls in calm markets and \pctAllBuyFall\,\% after a fall are unanimous Buys
(Table~\ref{tab:correlation}). Their errors are almost perfectly correlated ($\rho=\rhoErrAll$), so the three
models behave like $n_{\mathrm{eff}}=\nEffAll$ independent voters, and the majority is right \accMajAll\,\% of the
time, against \accJuryIndepAll\,\% if they were independent (Proposition~\ref{prop:jury}, Fig.~\ref{fig:jury}).

The second question is where the edge of \meTwo, the one source that sells often, comes from. The answer depends on
how it is scored. Per call, \meTwo{} has the lowest market beta of all sources (\betaMeTwo), its exits cannot be
told apart from random exits on the same dates ($p=\permP$), and its alpha is positive but not robustly so: the
block-bootstrap interval includes zero, the Newey--West interval does not. Scored as the dollar portfolio its trades imply, its invested dollars have a beta of \pfBetaAllMeTwo{} and an
alpha against NDX of \pfAlphaAllMeTwo\,\% a year ($t=\pfTAllMeTwo$), the same as its entry source \gtTwo; most of
it was earned on two stocks, MU and AMD.

\paragraph{Contributions}
\begin{enumerate}
\item \emph{Correlated bullish bias.} A live, by-regime measurement of how much three LLM trading signals
agree and err together, with the correlated-jury accuracy that their correlation implies (Section~\ref{sec:bias}).
\item \emph{All-Buy tracks the market.} The dominant outcome, all three saying Buy, is right only a few points
more often than the market goes up: \gapAllBuyCalm{} and \gapAllBuyFall{} points above the base rate in calm
markets and after a fall, before any exposure adjustment. A simple model shows that a strong shared bias drives this gap to zero
(Proposition~\ref{prop:allbuy}, Section~\ref{sec:game}).
\item \emph{An entry edge, not low exposure.} A per-call and dollar-portfolio analysis of \meTwo{} showing
that its edge comes from its \gtTwo{} entries, concentrated in two positions, and that its low per-call
beta comes from Sells scored as negative-beta calls, not from defensive holdings (Section~\ref{sec:me2}).
\item \emph{Per-call and portfolio scoring disagree.} Per call, whether the same policy's alpha excludes zero
depends on the interval method; as a dollar portfolio it is borderline significant; we report both for every source (Table~\ref{tab:portfolio}).
\item \emph{Robustness.} Costs, a deflated Sharpe ratio, factor baselines, a later entry price and an
independent price check (Section~\ref{sec:live}).
\item \emph{What transfers to production.} Position limits applied to every signal source, the stock-picking
test on a holdout, and DeepSeek's share of Sells as a market signal (Section~\ref{sec:production}).
\end{enumerate}
An exit counterfactual (Section~\ref{sec:dissent}) is reported as a preliminary observation: per lot, exits that
close lots often (the DeepSeek Sell or an aging exit) raise the win rate and the return per day held; at the
portfolio level no difference in alpha per invested dollar from the exits is detectable, though they limit the
capital at risk.

The results are from one market and one bear episode, and several are negative. Section~\ref{sec:discussion}
states what they do and do not show.

\begin{figure}[t]
\centering
\includegraphics[width=\columnwidth]{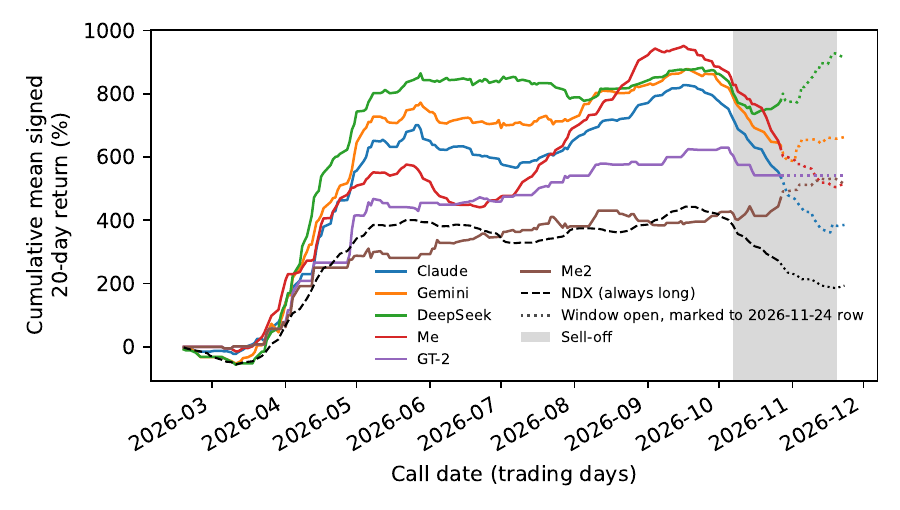}
\caption{Cumulative mean signed 20-day return of each source on a common grid of trading dates (a source adds its % num-ok: claim labels, horizon, confidence level or dates, not results
mean signed return on the dates it called and zero otherwise), against always-long NDX. Shaded: the
October--November 2026 sell-off. Dotted tails: calls whose 20-day window is % num-ok: horizon and calendar year in a caption
still open, marked to the 23 November close; they are shown for completeness and enter no table or number.} % num-ok: horizon and calendar year in a caption
\label{fig:cumulative}
\end{figure}

\section{Related work}\label{sec:related}

Table~\ref{tab:related} places this paper among the closest studies. Four strands meet here: correlated errors in
ensembles of LLMs, LLM trading systems and how they are evaluated, directional bias of LLMs in finance, and the
analysis of exits and market exposure.

\paragraph{Correlated errors in LLM ensembles}
Morante et al.\ note that combining the votes of several LLMs rests on the Condorcet jury theorem, which needs
independent voters, while models that share pre-training data, architectures, alignment or prompts tend to fail on
the same questions; the decisive quantity is the correlation of their errors \cite{morante2026correlation}. Kim
et al.\ measure this across more than three hundred LLMs and find substantial error correlation, highest among the
larger and more accurate models even across architectures and providers \cite{kim2025correlated}. In radiology,
a panel of multimodal LLMs from several developers was evaluated as a measuring instrument, with the independence
of its errors measured directly \cite{vakili2026correlated}. Closest to our setting, Chincholikar et al.\ vary
provider diversity in five-agent ensembles on a directional financial-analysis task and relate it to agreement
and cost \cite{chincholikar2026the}. A survey organises the methods for combining several LLMs
\cite{chen2025harnessing}, and a related line asks whether one LLM can stand in for the wisdom of a crowd of human
raters \cite{trott2023large}. On the theory side, jury theorems with correlated votes go back at least to Ladha
\cite{ladha1992the}, and recent work generalises the theorem to agents who are both unequally competent and
dependent \cite{karge2024to}. These studies use benchmarks, experiments or other domains. We add a measurement on
live trading calls, split by an ex-ante market regime, and report the direction of the shared error, which here is
bullish; Section~\ref{sec:method} uses a Beta-mixed model for three voters.

\paragraph{Consensus and herding among agents}
In medical question answering, agreement among models has been shown to be a weak signal of correctness
\cite{wang2026unanimously}, and LLMs can change their public decisions after observing ambiguous social signals
from other agents \cite{sheshanarayana2026phantom}. Wu et al.\ study the trade-off between consensus and diversity
in multi-agent systems and argue that explicit coordination, such as prompting or voting, risks premature
homogenisation \cite{wu2025the}. Multi-agent debates can converge to a collective bias, with a phase transition
once conformity passes a threshold \cite{okawa2026emergence}, and a taxonomy of failures of multi-agent LLM systems
shows how often their gains are small \cite{cemri2025why}. In finance, an event-replay stress test measures, in a
reduced-form sense, whether several LLM instances given similar prompts and market signals herd
\cite{apte2026do}, and after the public launch of ChatGPT the order flow in hard-to-read firms became more
one-sided, consistent with errors made more correlated by reliance on common models \cite{wang2026large}. For
human experts, herding makes probabilistic forecasts too similar in location \cite{jia2020herding}. Our setting has
no communication between the models at all: they are asked separately, and they still agree.

\paragraph{Game theory and LLMs}
Surveys map the use of game theory to study and to design LLM agents \cite{sun2025game,zhou2025exploring}. LLM
agents have been placed in a Keynesian beauty contest \cite{lu2025gametheory}, and a game-theoretic multi-agent
framework has been reported to lower hallucination rates of lightweight models on rule-constrained scientific
benchmarks \cite{liu2026game}. Earlier work by the author classifies the disagreement among several LLM trading
signals into six game types and finds that the type distinguishes directional accuracy; a human analyst who used
these classes to weight LLM disagreement had the best risk-adjusted profile in that study
\cite{zhang2026gametheoretic}. Here the same classes are used to show why they collapse: with a strong shared bias,
almost all three-way calls fall into the all-Buy class.

\paragraph{LLM trading systems and their evaluation}
Multi-agent trading frameworks give LLM agents the roles of a trading firm \cite{xiao2024tradingagents}, and a
survey reviews LLM agents in financial trading and notes that most evidence comes from backtests
\cite{ding2026large}. Backtests of LLM signals face two specific problems. The first is look-ahead bias: a model
may know what followed the news it is asked about \cite{glasserman2023assessing}, and this contamination can be
tested statistically \cite{gao2026detecting}. The second is selection: reported LLM advantages weaken over longer
periods and wider universes once survivorship and data snooping are controlled \cite{li2026can}, and a strategy
chosen as the best of many is inflated by the number of candidates tried \cite{olagoke2026trialsadjusted}. Live
designs avoid the first problem. A deployed multi-agent equity system has been validated on signals generated live
\cite{fatouros2026signal}, and ensembles of LLMs have been scored against live prediction markets
\cite{tsiklauri2026evaluating}. Our calls are also live; the \meTwo{} rules, however, were developed on the same
data, and we address selection only with a deflated Sharpe ratio and factor baselines.

\paragraph{Directional bias of LLMs in finance}
LLMs in investment analysis show built-in preferences when their parametric knowledge conflicts with market data
\cite{lee2025your}, LLM investment recommendations show systematic preferences for specific products
\cite{zhi2025exposing}, and Fin-Bias evaluates LLM financial decisions under human-like biases
\cite{hu2026finbias}; practitioner guidance now treats such biases as a governance risk \cite{tait2026managing}.
Frontier models show a disposition effect when asked which of two positions to sell \cite{pavlov2026naming}, and
LLM trading agents on Brazilian equities show provider-dependent behavioural asymmetries, with evidence strongest
for trading frequency \cite{cataneo2026providerdependent}. An earlier live benchmark of the same three LLMs by the
author found that no single model dominated and that the models differed far more in how often they bought than in
accuracy \cite{zhang2026benchmarking}. All of these find bias in prompts, in one-model agents or per model. We
measure something else: a shared bullish bias across three providers in live calls, and its cost by regime.

\paragraph{Buy and Sell recommendations}
The performance of analysts' Buy and Sell recommendations has long been studied separately
\cite{stickel1995the}. Analysts may issue optimistic public recommendations while passing more accurate
information to fund managers \cite{hirshleifer2022do}, and may issue Sells to manage how their recommendation
distribution looks \cite{kang2020out}. The LLM Sells in our data are similarly scarce (Table~\ref{tab:sources}),
but on their own they are not informative (Section~\ref{sec:me2}); at the portfolio level, exits on them add no detectable alpha per invested dollar but limit the capital at risk.

\paragraph{Exits, attribution and regimes}
An exit rule can be read as an option on the post-entry path, which splits a strategy's expectancy into parts
that do and do not depend on the exit \cite{tolusic2026every}. Agent Policy-Value Audit separates the composition
of an agent's trades from its event selection \cite{chen2026agent}. Regime-aware LLM trading systems detect a
regime and adapt risk control \cite{qiao2026drift} or reassign credit among specialist agents
\cite{pei2026market}. We do not add a regime detector; we explain the regime dependence of a two-sided,
multi-source policy by its market exposure, and test the exit's contribution with counterfactuals. For the
exposure analysis we use the standard five-factor and momentum models \cite{fama2015a,carhart1997on}.

\begin{table*}[t]
\centering
\caption{The closest studies and this paper. A mark means the study, as described in its abstract, has the
feature: $\checkmark$ yes, $\sim$ in part, -- no or not stated.}
\label{tab:related}
\begin{tabular}{lcccccc}
\toprule
Study & Live signals & Finance & Several providers & Error correlation & Direction of error & Exposure analysis \\
\midrule
Morante et al.\ \cite{morante2026correlation} & -- & -- & $\checkmark$ & $\checkmark$ & -- & -- \\
Kim et al.\ \cite{kim2025correlated} & -- & -- & $\checkmark$ & $\checkmark$ & -- & -- \\
Chincholikar et al.\ \cite{chincholikar2026the} & -- & $\checkmark$ & $\checkmark$ & $\checkmark$ & -- & -- \\
Apte et al.\ \cite{apte2026do} & -- & $\checkmark$ & -- & $\sim$ & -- & -- \\
Fatouros et al.\ \cite{fatouros2026signal} & $\checkmark$ & $\checkmark$ & -- & -- & -- & $\sim$ \\
FINSABER \cite{li2026can} & -- & $\checkmark$ & -- & -- & -- & -- \\
Zhang \cite{zhang2026benchmarking} & $\checkmark$ & $\checkmark$ & $\checkmark$ & $\sim$ & $\sim$ & -- \\
This paper & $\checkmark$ & $\checkmark$ & $\checkmark$ & $\checkmark$ & $\checkmark$ & $\checkmark$ \\
\bottomrule
\end{tabular}
\end{table*}

\section{Data and system}\label{sec:data}

\paragraph{Ledger}
The data are a ledger of daily Buy and Sell calls on \nTickers{} large US-listed stocks, mostly technology (\nTickersStart{} from the first day; the others
were added during the sample, the last in July), from \sampleStart{} to \sampleEnd{}, kept by the author's
trading system, with the NDX index as the market. The daily collection of LLM calls on US technology stocks was
introduced in an earlier live benchmark by the author \cite{zhang2026benchmarking}. Each row carries the date, the ticker, the source, the signal and
the ledger price. The analysis ends with the ledger row of 24 November 2026, whose price is the 23 November close; later rows are ignored, so every number can be rebuilt. Claude's calls for 23 and 24 November were produced after the fact, from the same day's inputs, because its API failed on those days; on those two days \gtTwo{} and \meTwo{} acted without them. % num-ok: claim labels, horizon, confidence level or dates, not results
% num-ok: sample end date
The \gtTwo{} and \meTwo{} records in the ledger follow their current rules throughout the sample; the
aging exit used before 26 July does not appear in the ledger. Calls on the NDX index itself (\nNdxCalls{} in the ledger) are not counted as stock calls, and
one ledger price, AMD on 24 February, is dropped as an error (140.61 between 199.90 and 194.90). The ticker
list is not survivorship-free (Section~\ref{sec:discussion}). % num-ok: the bad price itself

\begin{figure}[t]
\centering
\includegraphics[width=0.92\textwidth]{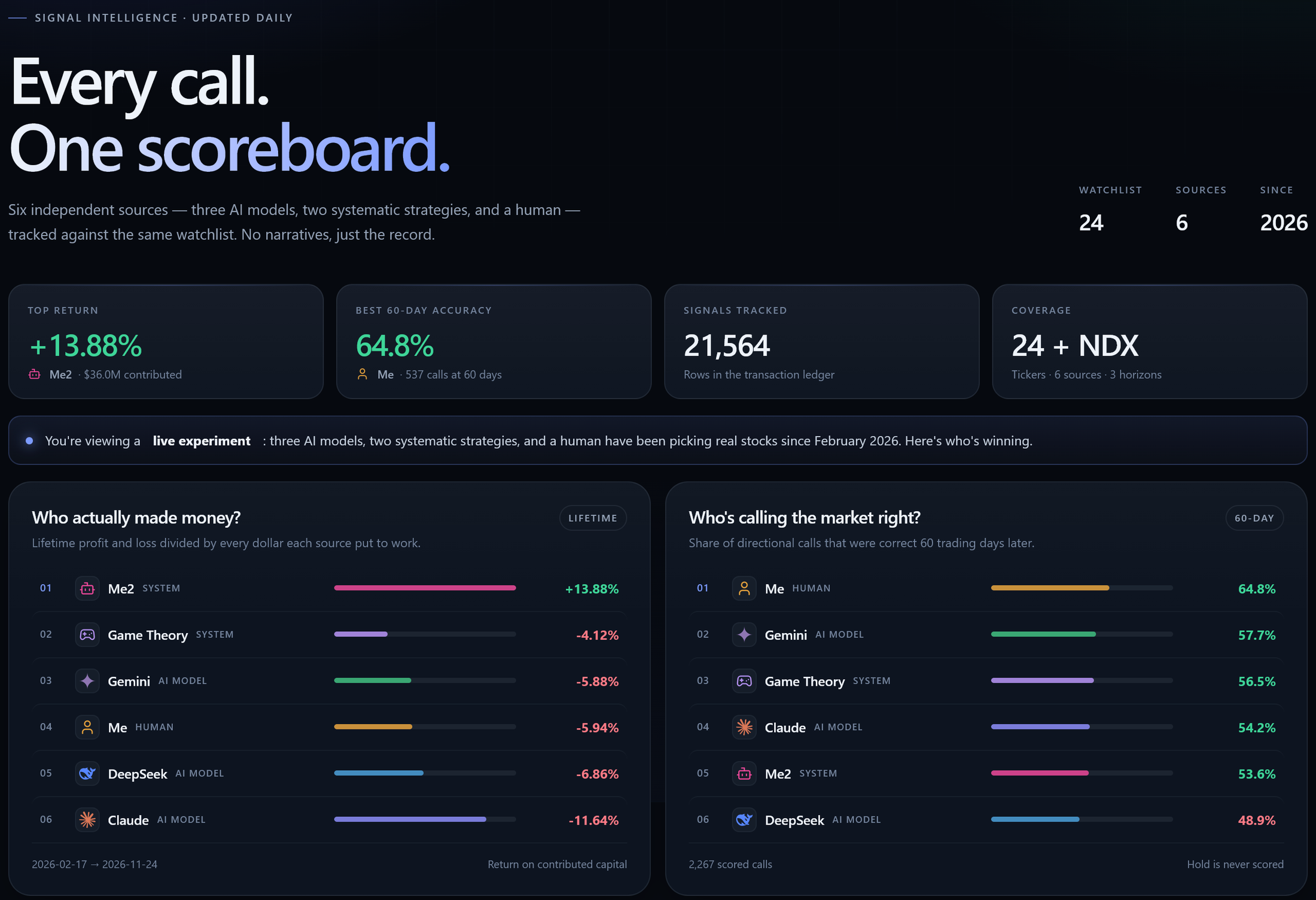}
\caption{The Manual\_Stock dashboard (overview) at the end of the sample: the six sources and the two headline
scoreboards. Screenshot of the live system; its measures differ from the paper's.}
\label{fig:dashoverview}
\end{figure}

\begin{figure}[t]
\centering
\includegraphics[width=0.92\textwidth]{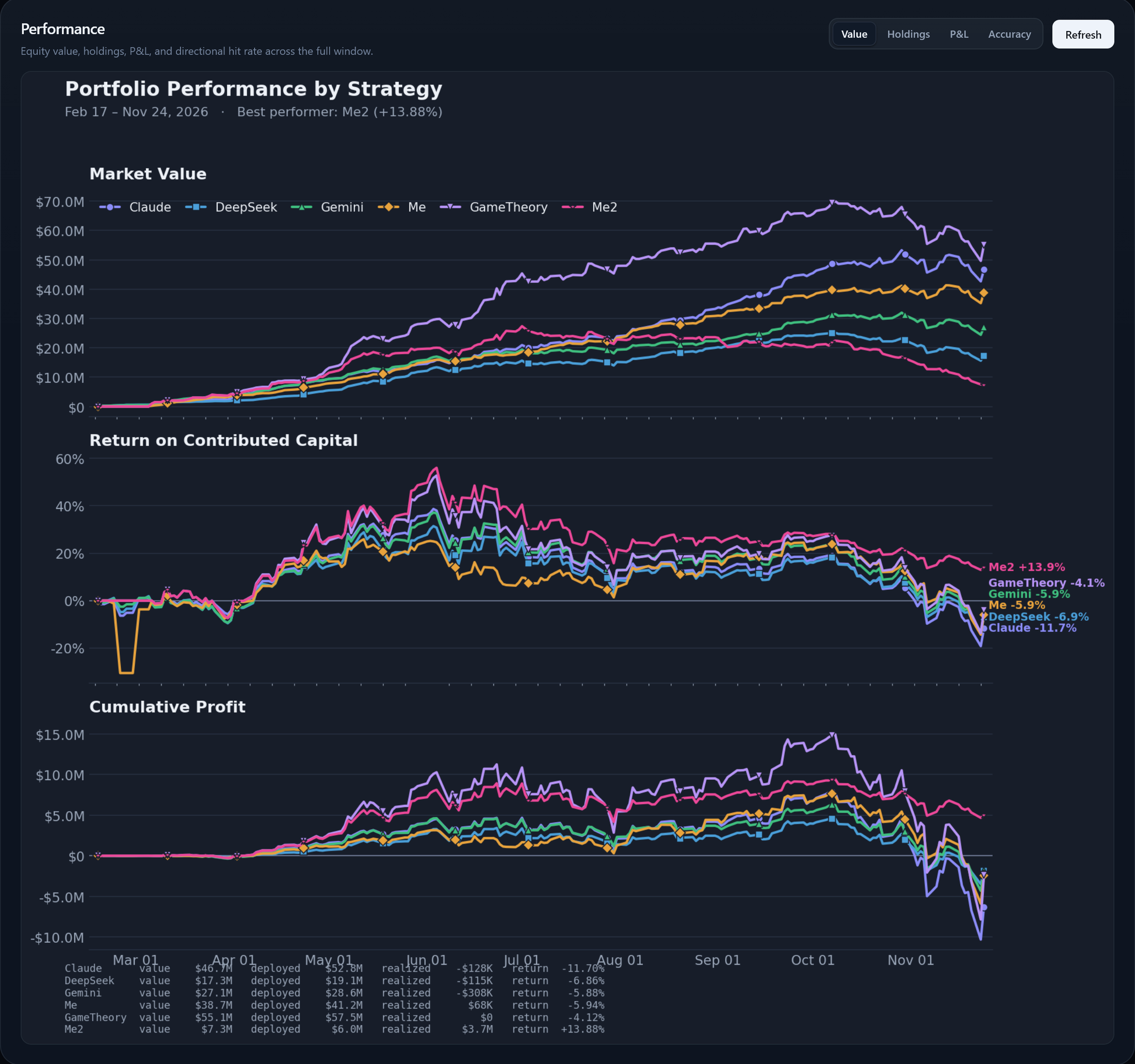}
\caption{The dashboard's performance panel: market value, return on contributed capital and cumulative profit of
each source over the whole period, ending in the October--November sell-off. Screenshot of the live system.}
\label{fig:dashperformance}
\end{figure}

\begin{figure}[t]
\centering
\includegraphics[width=0.92\textwidth]{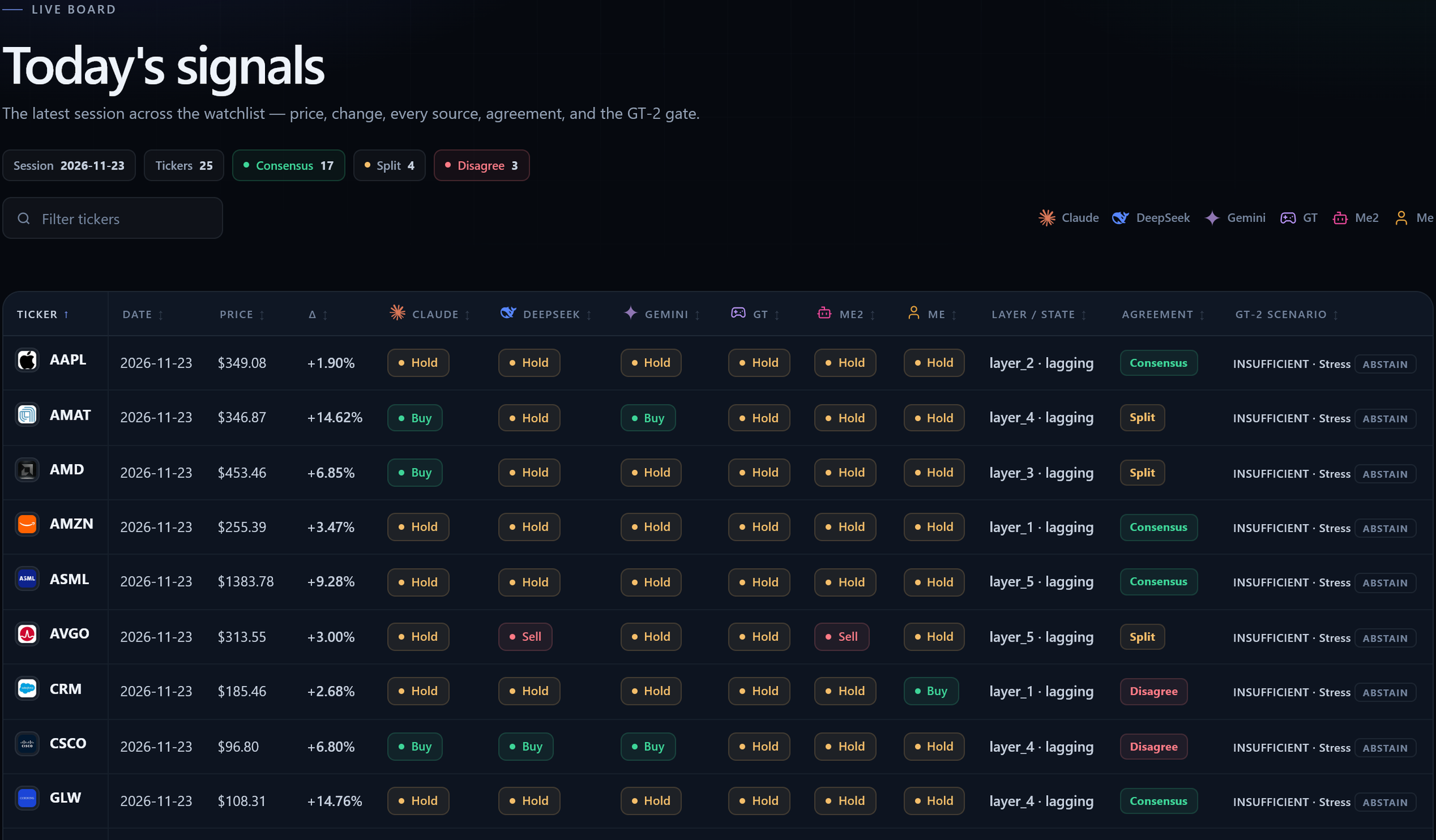}
\caption{The dashboard's board for the last session of the sample (first rows): price and change, each source's
signal, the agreement among sources, and the game type and decision of \gtTwo. Screenshot of the live system.}
\label{fig:dashtoday}
\end{figure}

\paragraph{The live system}
The ledger is produced by the author's Manual\_Stock system, which publishes every call on a web dashboard
(Data availability). Figs.~\ref{fig:dashoverview}--\ref{fig:dashtoday} show it as it stood at the end of the
sample: the headline scoreboard, the performance of each source over the whole period, and the board of the last
session, with each source's signal per ticker, their agreement and the game type that \gtTwo{} assigns. The
dashboard uses its own measures, return on contributed capital and accuracy at a longer horizon, so its numbers
are not the paper's and are shown only to document the system; every number in the text comes from the analysis
scripts. Holds, which the dashboard displays, are not calls in this paper.

\paragraph{Sources}
Table~\ref{tab:sources} lists the six sources. Claude, Gemini and DeepSeek are three LLMs asked for a daily call
on each ticker. Me is the author's manual call. \gtTwo{} combines the three LLM calls with a game-theoretic rule
built on the author's earlier work on game-theoretic readings of LLM disagreement \cite{zhang2026gametheoretic}. \meTwo{} is the role-split policy of Section~\ref{sec:me2policy}: it buys on a \gtTwo{}
Buy, sells on a DeepSeek Sell, and holds at most five lots per ticker. Its rules were developed on this
ledger; the current version dates from 26 July 2026, when an aging exit was removed. % num-ok: rule-change date
\meTwo{} made \cfLedgerBuys{} Buys in the ledger.

\paragraph{Prices}
An independent check against daily closes from another source shows that the ledger price dated $d$ is the close
of the previous trading day (exact on \exactLagNvda\,\% of NVDA days after a one-day shift,
Table~\ref{tab:pricecheck}). Returns therefore start at the last close known when the call was made. The same check
shows that the October--November 2026 sell-off was deeper than the ledger suggests: from 7 October to 20 November % num-ok: sell-off dates
NVDA fell \trueSelloffNvda\,\% and NDX \trueSelloffNdx\,\% on independent closes, against \ledgerSelloffNvda\,\% and
\ledgerSelloffNdx\,\% on ledger prices dated the same days.

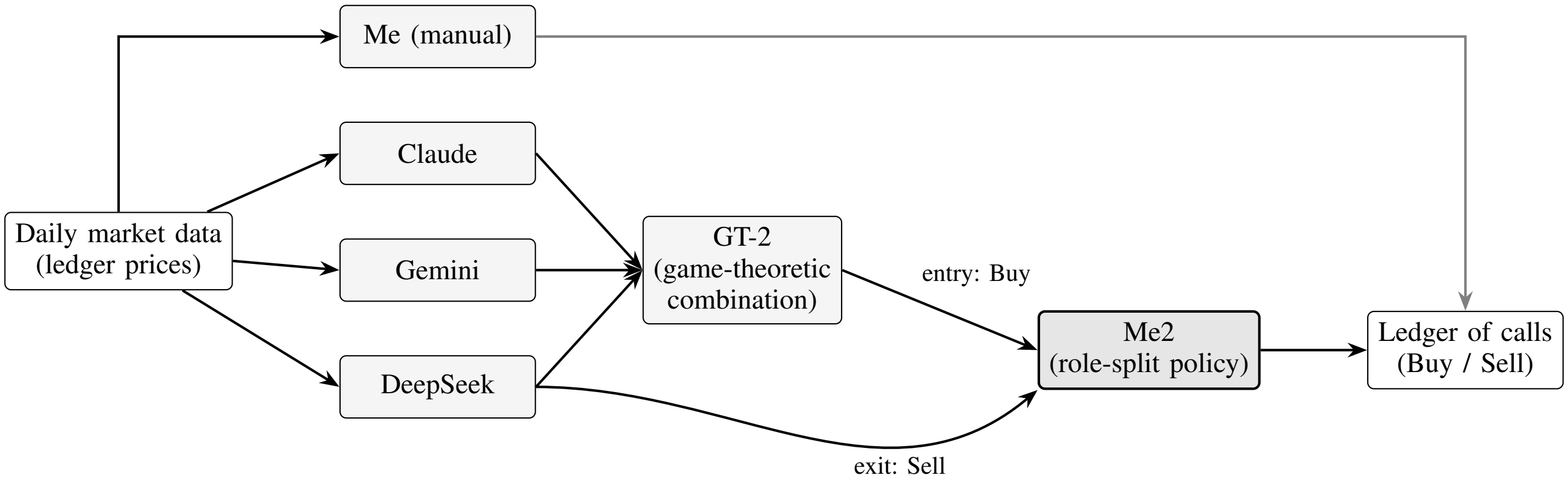
\begin{figure}[t]
\centering
\begin{tikzpicture}[
  node distance=6mm and 12mm, % num-ok: tikz layout, not a result
  box/.style={draw, rounded corners=2pt, minimum height=7mm, minimum width=22mm, align=center, font=\small}, % num-ok: tikz layout, not a result
  src/.style={box, fill=black!4}, % num-ok: tikz layout, not a result
  pol/.style={box, fill=black!10, line width=0.8pt}, % num-ok: tikz layout, not a result
  arr/.style={-{Stealth[length=2.2mm]}, thick}] % num-ok: tikz layout, not a result
\node[box] (data) {Daily market data\\(ledger prices)};
\node[src, right=of data, yshift=11mm] (cl) {Claude}; % num-ok: tikz layout, not a result
\node[src, below=of cl] (ge) {Gemini};
\node[src, below=of ge] (ds) {DeepSeek};
\node[src, above=of cl] (me) {Me (manual)};
\node[src, right=of ge] (gt) {\gtTwo\\(game-theoretic\\combination)};
\node[pol, right=22mm of gt, yshift=-9mm] (me2) {\meTwo\\(role-split policy)}; % num-ok: tikz layout, not a result
\node[box, right=of me2] (led) {Ledger of calls\\(Buy / Sell)}; % num-ok: tikz layout, not a result
\draw[arr] (data) -- (cl.west);
\draw[arr] (data) -- (ge.west);
\draw[arr] (data) -- (ds.west);
\draw[arr] (data) |- (me.west);
\draw[arr] (cl.east) -- (gt.west);
\draw[arr] (ge.east) -- (gt.west);
\draw[arr] (ds.east) -- (gt.west);
\draw[arr] (gt.east) -- node[above right, font=\footnotesize, pos=0.35] {entry: Buy} (me2.west); % num-ok: tikz layout, not a result
\draw[arr] (ds.east) to[out=0, in=220] node[below, font=\footnotesize, pos=0.7] {exit: Sell} (me2.south west); % num-ok: tikz layout, not a result
\draw[arr] (me2) -- (led); % num-ok: tikz layout, not a result
\draw[arr, black!50] (me.east) -| (led.north); % num-ok: tikz layout, not a result
\end{tikzpicture}
\caption{The six sources. Three LLMs and the author give daily calls; \gtTwo{} combines the three LLM calls; \meTwo{}
enters on a \gtTwo{} Buy and exits on a DeepSeek Sell, with a cap on open lots. All calls go to one ledger.}
\label{fig:system}
\end{figure}

\begin{figure}[t]
\centering
\includegraphics[width=0.8\columnwidth]{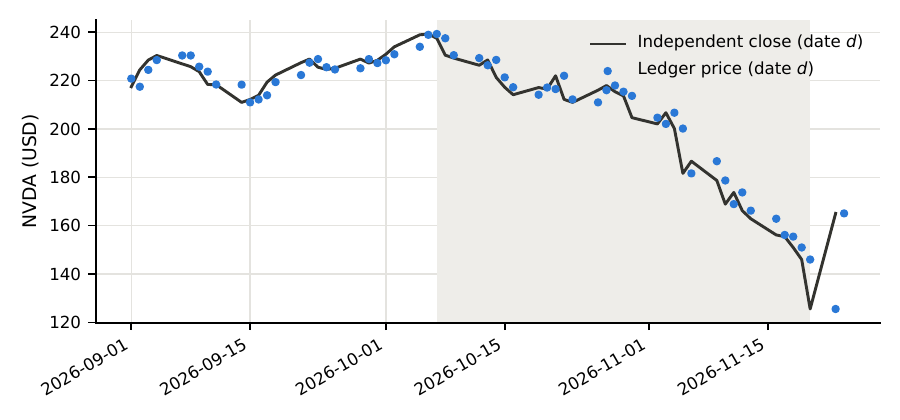}
\caption{NVDA: independent daily closes (line) and ledger prices (points) on the same dates, from September to the
end of the sample. Shaded: the sell-off window. The ledger points sit one trading day behind the line.}
\label{fig:pricecheck}
\end{figure}

Fig.~\ref{fig:system} shows how the six sources are produced and how the role-split policy uses them, and
Table~\ref{tab:summary} describes the distribution of signed returns per source. The distributions are wide and
right-skewed for every source, which is one reason the paper reports intervals from a block bootstrap rather than
from normal theory. Fig.~\ref{fig:pricecheck} shows the price check on NVDA over the sell-off: the ledger points
track the independent closes of the day before.

\begin{table}[t]
\centering
\caption{Sources: calls, Buys and Sells.}
\label{tab:sources}
\begin{tabular}{lrrrr}
\toprule
Source & Calls & Buy & Sell & Sell (\%) \\
\midrule
Claude & 1109 & 1089 & 20 & 1.8 \\
Gemini & 685 & 627 & 58 & 8.5 \\
DeepSeek & 746 & 541 & 205 & 27.5 \\
Me & 833 & 830 & 3 & 0.4 \\
GT-2 & 115 & 115 & 0 & 0.0 \\
Me2 & 145 & 72 & 73 & 50.3 \\
\bottomrule
\end{tabular}

\end{table}

\begin{table*}[t]
\centering
\caption{Distribution of signed 20-day returns per source (\%): calls, distinct call dates, mean, standard % num-ok: horizon in a caption
deviation, 5th and 95th percentiles and median, skewness, excess kurtosis and share of positive calls. % num-ok: percentile definition, not a result
Descriptive; windows overlap.}
\label{tab:summary}
\begin{tabular}{lrrrrrrrrrr}
\toprule
Source & $n$ & Dates & Mean & s.d. & P5 & Median & P95 & Skew. & Ex.\ kurt. & $>0$ (\%) \\
\midrule
Claude & 930 & 160 & +1.83 & 17.49 & -24.0 & +1.1 & +32.2 & +1.13 & 3.83 & 54.0 \\
Gemini & 603 & 160 & +3.15 & 18.33 & -20.3 & +1.0 & +36.9 & +1.41 & 4.05 & 53.6 \\
DeepSeek & 641 & 164 & +3.15 & 17.98 & -21.8 & +1.7 & +34.5 & +1.03 & 3.76 & 56.0 \\
Me & 688 & 158 & +2.59 & 17.73 & -23.8 & +2.3 & +30.0 & -0.05 & 4.61 & 56.2 \\
GT-2 & 115 & 72 & +8.06 & 20.79 & -14.8 & +3.6 & +54.3 & +1.85 & 3.91 & 61.7 \\
Me2 & 125 & 85 & +5.76 & 17.31 & -15.7 & +4.2 & +36.0 & +1.59 & 5.22 & 64.8 \\
\bottomrule
\end{tabular}

\end{table*}

% S6 owns this file. Definitions, the Me2 policy, the estimators, and the two propositions with proofs.
\section{Method}\label{sec:method}

\ifdefined\definition\else\newtheorem{definition}{Definition}\fi
\ifdefined\proposition\else\newtheorem{proposition}{Proposition}\fi

\subsection{Calls, returns and regimes}

\begin{definition}[Call and signed return]\label{def:call}
A \emph{call} is a tuple $(d,i,s,g)$: on ledger date $d$, source $s$ issues signal $g\in\{+1,-1\}$ (Buy, Sell)
for ticker $i$. The ledger has one row per date on which the signal job ran: every trading day, plus a few
weekend and holiday dates on which the price repeats the last close. Let $P_{i}(d)$ be the ledger price of $i$
dated $d$ and $d+H$ the date $H$ ledger rows of $i$ later, with horizon $H=20$ rows (about one trading month). % num-ok: names (Me2, GT-2), dates, or method parameters, not results
Here $d+H$ denotes the date of that row, not calendar arithmetic. The forward return is
$R_i(d)=P_i(d+H)/P_i(d)-1$ and the \emph{signed return}
of the call is $y=g\,R_i(d)$. The call is \emph{correct} when $y>0$; a zero return counts as a miss. The market
return $m(d)$ is the forward return of NDX over its own $H$ ledger rows starting at the same date $d$. The NDX
ledger has no row on a few trading days that the stocks have, so over those gaps an NDX window covers a day or two
more than the stock's window, and a call dated on one of those days has no market return and is dropped.
\end{definition}

The ledger price dated $d$ is the previous trading day's close (see the independent price check below), so
$R_i(d)$ starts at the last close known when the call was made. All windows below are counted in ledger rows with
this same convention. As a robustness check every main estimate is
recomputed with entry one ledger row later. When $d$ is a trading day this price is the close of the signal day
itself; when $d$ is a weekend or holiday row it is the last close before the next ledger date.

\begin{definition}[Ex-ante regime]\label{def:regime}
Date $d$ is \emph{after a fall} when the trailing NDX return over the past $H$ rows, $P_{\mathrm{NDX}}(d)/
P_{\mathrm{NDX}}(d-H)-1$, is negative, and \emph{calm} otherwise. Both prices follow the ledger convention, so the window ends at the
close before $d$ and the regime uses only data known at $d$ and is the
same for every source on that date. The \emph{base rate} of a regime is the share of positive $R_i(d)$ over all
stock-days in it (a zero return counts as not positive, as for calls), over the same window as the calls; it is
a descriptive benchmark, not a forecast.
\end{definition}

All results use the whole sample, except that Section~\ref{sec:production} also reports a holdout.

\subsection{Correlated voters}

For the three LLMs $j\in\{1,2,3\}$ (Claude, Gemini, DeepSeek), restrict to ticker-dates where all three call and
let $X_j=1$ if LLM $j$'s call is correct and $0$ otherwise. The error indicator is $1-X_j$, so the error correlation
equals the correlation of the $X_j$. We write $p$ for the mean of the three accuracies $\Pr(X_j=1)$ and $\rho$ for the mean pairwise
correlation of the $X_j$, estimated per regime, and report Cohen's $\kappa$ of the signals as a second agreement
measure. The \emph{effective number of independent voters} is the design-effect count
\begin{equation}\label{eq:neff}
n_{\mathrm{eff}}=\frac{3}{1+2\rho},
\end{equation}
which equals $3$ when $\rho=0$ and $1$ when $\rho=1$.

\begin{definition}[Correlated jury]\label{def:jury}
The calls are an exchangeable correlated jury with parameters $(p,\rho)$, $0<\rho<1$, when there is a latent
$\pi\sim\mathrm{Beta}(a,b)$ with $a=p(1-\rho)/\rho$ and $b=(1-p)(1-\rho)/\rho$, and given $\pi$ the $X_j$ are
independent $\mathrm{Bernoulli}(\pi)$. Then $\Pr(X_j=1)=p$ and $\mathrm{corr}(X_j,X_k)=1/(a+b+1)=\rho$ for $j\neq k$.
The cases $\rho=0$ (independent voters) and $\rho=1$ (identical voters) are the limits.
\end{definition}

The definition imposes equal accuracies and equal pairwise correlations; the data give three accuracies and three
correlations, which we summarise by their means.

Pairwise moments alone do not fix the majority accuracy of three voters, because the probability that all three
are correct is a third-order moment. Definition~\ref{def:jury} is the standard one-parameter way to close the
model; it is the model behind the model curve in Table~\ref{tab:correlation} and Fig.~\ref{fig:jury}.

\begin{proposition}[Majority accuracy falls with error correlation]\label{prop:jury}
Under Definition~\ref{def:jury} with $1/2<p<1$, the probability that the majority of three voters is correct is
\begin{equation}\label{eq:maj}
M(\rho)=\frac{p\,\bigl(p+\rho(1-p)\bigr)\bigl(3-2p+\rho(2p-1)\bigr)}{1+\rho}.
\end{equation}
$M$ is strictly decreasing on $[0,1]$, with $M(0)=3p^2-2p^3$ (Condorcet's independent value) and $M(1)=p$.
\end{proposition}

\begin{proof}
Let $S=X_1+X_2+X_3$. Given $\pi$, $\Pr(S\ge 2\mid\pi)=3\pi^2(1-\pi)+\pi^3=3\pi^2-2\pi^3$, so
$M=3\,\mathbb{E}[\pi^2]-2\,\mathbb{E}[\pi^3]$. With $c=a+b=(1-\rho)/\rho$, the Beta moments are
$\mathbb{E}[\pi^2]=\frac{a(a+1)}{c(c+1)}$ and $\mathbb{E}[\pi^3]=\mathbb{E}[\pi^2]\frac{a+2}{c+2}$. Since $a=pc$,
$c+1=1/\rho$ and $c+2=(1+\rho)/\rho$,
\[
\mathbb{E}[\pi^2]=p\,(pc+1)\rho=p\bigl(p+\rho(1-p)\bigr),\qquad
\frac{a+2}{c+2}=\frac{p(1-\rho)+2\rho}{1+\rho}=\frac{p+\rho(2-p)}{1+\rho}.
\]
Hence
\[
M=\mathbb{E}[\pi^2]\Bigl(3-\frac{2p+2\rho(2-p)}{1+\rho}\Bigr)
 =\mathbb{E}[\pi^2]\,\frac{3-2p+\rho(2p-1)}{1+\rho},
\]
which is \eqref{eq:maj}. The formula is continuous at $\rho=0$ and $\rho=1$, where it gives $p^2(3-2p)$ and
$p\cdot 1\cdot 2/2=p$, the values for independent and for identical voters.

For monotonicity write $M=p\,(A+B\rho)(C+D\rho)/(1+\rho)$ with $A=p$, $B=1-p$, $C=3-2p$, $D=2p-1$. Then
$M'(\rho)=p\,N(\rho)/(1+\rho)^2$, with $p/(1+\rho)^2>0$, where
\[
N(\rho)=\bigl(B(C+D\rho)+D(A+B\rho)\bigr)(1+\rho)-(A+B\rho)(C+D\rho)
=(BC+AD-AC)+2BD\rho+BD\rho^2 .
\]
Here $BC+AD-AC=(1-p)(3-2p)+p(2p-1)-p(3-2p)=3-9p+6p^2=3(1-p)(1-2p)$ and $BD=(1-p)(2p-1)$, so
\[
N(\rho)=(1-p)(2p-1)\bigl(\rho^2+2\rho-3\bigr)=(1-p)(2p-1)(\rho+3)(\rho-1).
\]
For $1/2<p<1$ the first two factors are positive, $\rho+3>0$, and $\rho-1<0$ on $[0,1)$. So $M'<0$ on $(0,1)$.
Since $M$ is continuous on $[0,1]$, the mean value theorem gives $M(\rho_1)>M(\rho_2)$ for all
$0\le\rho_1<\rho_2\le 1$, so $M$ is strictly decreasing on the closed interval.
\end{proof}

Proposition~\ref{prop:jury} is what Table~\ref{tab:correlation} checks: with the measured $p$ and $\rho$, the
independent value $M(0)$ is \accJuryIndepAll\,\% and the observed majority accuracy is \accMajAll\,\%, at
$\rho=\rhoErrAll$ and $n_{\mathrm{eff}}=\nEffAll$. The assumptions (equal $p$ and $\rho$ across voters, a
Beta-mixed common accuracy) are not tested; the proposition gives the direction of the effect, not a fit.

\subsection{A model of the shared bullish bias}\label{sec:model}

The game of Section~\ref{sec:game} is an illustration. We state it precisely so that one limit can be proved.
Let $r$ be a stock's forward return with $\mathbb{E}|r|<\infty$ and $\Pr(r>0)=u$ (the base rate). Each LLM
$j$ calls
\begin{equation}\label{eq:model}
g_j=\operatorname{sign}\bigl(z+b+e_j\bigr),
\end{equation}
where $z$ is a real-valued signal the three models share (it may be correlated with $r$), $b\ge 0$ is a common
bullish bias, and $e_j$ are real-valued noise terms with any joint law (correlated or not). Ties
($z+b+e_j=0$) count as Sell. The disagreement classes of the author's earlier work are the outcomes
of \eqref{eq:model}; write $\mathrm{AB}_b$ for the event that all three call Buy.

\begin{proposition}[All-Buy accuracy tends to the base rate]\label{prop:allbuy}
In model \eqref{eq:model}, as $b\to\infty$:
(i) $\Pr(\mathrm{AB}_b)\to 1$;
(ii) $\Pr(r>0\mid \mathrm{AB}_b)\to u$, so the all-Buy gap over the base rate tends to $0$; and
(iii) $\mathbb{E}[r\mid\mathrm{AB}_b]\to\mathbb{E}[r]$. If moreover $r=\beta m+\varepsilon$ with
$\mathbb{E}|m|,\mathbb{E}|\varepsilon|<\infty$, the limit is $\beta\,\mathbb{E}[m]+\mathbb{E}[\varepsilon]$.
Here $\varepsilon$ is any integrable remainder (it may have a nonzero mean, which then plays the role of an
intercept). This holds for every joint law of $(z,e_1,e_2,e_3,r)$, whatever the noise correlation.
\end{proposition}

\begin{proof}
(i) Call $j$ is Sell only if $z+e_j\le -b$. Since $z+e_j$ is a real random variable,
$\Pr(z+e_j\le -b)\to 0$ as $b\to\infty$. By the union bound
$\Pr(\mathrm{AB}_b^{c})\le\sum_{j=1}^{3}\Pr(z+e_j\le -b)\to 0$.

(ii) For $b$ large enough that $\Pr(\mathrm{AB}_b)>0$,
\[
u-\Pr(\mathrm{AB}_b^{c})\le\Pr(r>0,\mathrm{AB}_b)\le u,
\]
so $\Pr(r>0\mid\mathrm{AB}_b)=\Pr(r>0,\mathrm{AB}_b)/\Pr(\mathrm{AB}_b)$ lies between
$(u-\Pr(\mathrm{AB}_b^{c}))/\Pr(\mathrm{AB}_b)$ and $u/\Pr(\mathrm{AB}_b)$. Both bounds tend to $u$ by (i).

(iii) $\mathbb{E}[r\,\mathbf{1}_{\mathrm{AB}_b}]=\mathbb{E}[r]-\mathbb{E}[r\,\mathbf{1}_{\mathrm{AB}_b^{c}}]$ and
$|\mathbb{E}[r\,\mathbf{1}_{\mathrm{AB}_b^{c}}]|\le\mathbb{E}[|r|\,\mathbf{1}_{\mathrm{AB}_b^{c}}]$. Because $|r|$ is
integrable and $\Pr(\mathrm{AB}_b^{c})\to 0$, absolute continuity of the integral gives
$\mathbb{E}[|r|\,\mathbf{1}_{\mathrm{AB}_b^{c}}]\to 0$. Dividing by $\Pr(\mathrm{AB}_b)\to 1$ gives the limit
$\mathbb{E}[r]$, and linearity of expectation gives $\beta\,\mathbb{E}[m]+\mathbb{E}[\varepsilon]$.
\end{proof}

Proposition~\ref{prop:allbuy} is a limit statement. It says that a strong enough shared bias makes the
all-Buy outcome carry the market's direction and nothing else; it does not say how large $b$ is for the three
LLMs, and it does not predict the size of a finite-$b$ gap. The measured gaps (\gapAllBuyCalm{} points calm,
\gapAllBuyFall{} after a fall, Table~\ref{tab:classes}) are reported as data, not as a test of the proposition.

\subsection{The Me2 policy}\label{sec:me2policy} % num-ok: names (Me2, GT-2), dates, or method parameters, not results

Me2 is a role-split policy over the other sources. Its current rules, in force since 2026-07-26, are applied every ledger date % num-ok: names (Me2, GT-2), dates, or method parameters, not results
$d$ and every ticker $i$, exits first:
\begin{enumerate}
\item \emph{Exit.} If Me2 holds lots of $i$ and DeepSeek calls Sell on $(d,i)$, one lot's worth of $i$ is sold, oldest lot first. % num-ok: names (Me2, GT-2), dates, or method parameters, not results
\item \emph{Entry.} If GT-2 calls Buy on $(d,i)$ and fewer than $5$ lots of $i$ are open, one equal-weight lot % num-ok: names (Me2, GT-2), dates, or method parameters, not results
is bought.
\end{enumerate}
The ledger's Me2 record follows these rules throughout the sample. Because exits run % num-ok: names (Me2, GT-2), dates, or method parameters, not results
first, a lot is never bought and sold on the same date. As a call source, Me2's entries are Buy calls and its % num-ok: names (Me2, GT-2), dates, or method parameters, not results
exits are Sell calls, scored like any other call (Definition~\ref{def:call}).

\paragraph{Exit counterfactuals}
To isolate the exit, a simplified policy that sells all lots of a ticker on an exit (the ledger sells one lot
per Sell) is run on the ledger signals with the entry rule and the lot cap fixed and the
exit swapped: the DeepSeek Sell, the Claude Sell, the Gemini Sell, a Sell from any LLM, no exit
(hold to the end of the sample), and the earlier aging exit with and without the DeepSeek Sell. A lot's return
runs from its entry price to its exit price; lots still open at the end are marked at the last price, which affects
mostly the no-exit variant. The lot's market return is the NDX return over the same dates. Of the
\cfReplayBuys{} replayed entries, \cfLedgerMatch{} match the \cfLedgerBuys{} Me2 Buys in the ledger. % num-ok: names (Me2, GT-2), dates, or method parameters, not results

\subsection{Exposure and alpha}

For each source, regress signed returns on the market return over all its scored calls,
\begin{equation}\label{eq:capm}
y_k=\alpha+\beta\,m_k+\epsilon_k .
\end{equation}
For counterfactual lots, $y_k$ and $m_k$ are the lot and market returns over the holding period. Two sources of
dependence matter: many calls share a date, and $H$-row windows starting on nearby dates overlap. We therefore
use a date-clustered Newey--West (Driscoll--Kraay) covariance. Let $X$ be the call-level design matrix with rows
$x_k^{\top}=(1,m_k)$ and $\hat\epsilon_k$ the OLS residuals; the scores $x_k\hat\epsilon_k$ are summed over the calls
of each date $t=1,\dots,T_d$ to $u_t$, with dates in order and $u_t=0$ for $t<1$ and $t>T_d$, and
\[
\hat S=\sum_t u_tu_t^{\top}+\sum_{\ell=1}^{L}\Bigl(1-\frac{\ell}{L+1}\Bigr)\sum_t\bigl(u_tu_{t-\ell}^{\top}+
u_{t-\ell}u_t^{\top}\bigr),\qquad \widehat{\mathrm{Var}}(\hat\theta)=(X^{\top}X)^{-1}\hat S(X^{\top}X)^{-1},
\]
with lag $L=H-1$ dates, the overlap of windows that start one row apart. The 95\,\% confidence interval for $\alpha$ is a percentile % num-ok: names (Me2, GT-2), dates, or method parameters, not results
interval from a moving-block bootstrap over dates (blocks of $H$ consecutive dates, all calls of a date resampled
together, 2000 draws, seed 20261120). % num-ok: names (Me2, GT-2), dates, or method parameters, not results

\paragraph{Factor-adjusted alpha}
To compare with standard factors, \eqref{eq:capm} is extended to the five Fama--French factors and momentum. For a
call on $(d,i)$, each factor's return is compounded over the trading days among the call's $H$ ledger rows.
A weekend or holiday row repeats the previous price, so the call's own return over those rows equals its return
over the trading days alone, and the two sides of the regression cover the same days. The dependent variable is the signed return
minus the compounded T-bill rate. Only calls whose window ends by the last factor date (\factorEnd) enter. The
covariance is the same date-clustered Newey--West estimator.

\paragraph{Entry and exit decomposition}
Using each group's own $\hat\beta$, the mean signed return of Me2's entries and of its exits are each split into a % num-ok: names (Me2, GT-2), dates, or method parameters, not results
market part, $\hat\beta$ times the mean market return over those calls, and a selection part, the remainder. The
split is model-based and descriptive, and has no confidence interval of its own.

\paragraph{Dollar-portfolio alpha}
Scoring calls one by one gives every call the same weight and ignores how long a position is held and how much
money it ties up. We therefore also score each source as the dollar portfolio its ledger trades imply. The BUY and
SELL rows (shares at the ledger price) give a share position per ticker. The dollar P\&L on ledger date $t$ is the
sum over tickers of the shares held at $t-1$ times the price change from $t-1$ to $t$, with a missing price carried
forward; the portfolio return $r_t$ is that P\&L divided by the dollars invested at $t-1$, on dates with money
invested. On ledger dates with no NDX row, the P\&L is added to the next NDX date. We regress
$r_t=\alpha+\beta\,m_t+\epsilon_t$ on the daily NDX return $m_t$ from the same ledger rows, so both sides share the
one-day lag of ledger prices, with Newey--West standard errors (lag \pfLag{} days), and report $\alpha$ annualised
(times 252). $\beta$ is then the exposure of the dollars actually invested; days in cash do not enter. The % num-ok: trading days per year
regression is run on all dates, and for \meTwo{} also without its two
largest contributors. No deflation for the number of sources or splits is applied to these $t$-statistics.

\subsection{Tests}

\paragraph{Exit permutation}
Do the exits pick stocks that fall? The null is exiting at random on the same dates. For each Me2 exit on $(d,i)$, % num-ok: names (Me2, GT-2), dates, or method parameters, not results
draw a ticker uniformly from the stocks with a known forward return on $d$, the exited stock included, and score
the draw as a fall when its forward return $R(d)$ over the next $H$ rows is negative. The statistic is the share
of falls. Its one-sided $p$-value is the share of 5000 random draws, each over all exits, whose statistic is at % num-ok: names (Me2, GT-2), dates, or method parameters, not results
least the observed one.

\paragraph{Placebo entries}
Do the entries pick the right stocks? The ledger's Me2 trades are split into first-in-first-out lots, each with a % num-ok: names (Me2, GT-2), dates, or method parameters, not results
buy date, a dollar amount and the dates and fractions of the Sells that close it. A placebo keeps all of these and
changes only the ticker. In the main version every real ticker is mapped to one distinct random ticker with a price
on its first buy date, so all lots of a ticker move together and the concentration of the real portfolio is kept;
a second version draws a ticker per lot. Each placebo portfolio is scored like the real one (daily alpha against
NDX and dollar P\&L, Section~\ref{sec:method}), over \plDraws{} draws; the one-sided $p$-value is the share of
draws at least as good as the real portfolio. The same test is repeated on the lots bought after 2026-07-26, % num-ok: names (Me2, GT-2), dates, or method parameters, not results
scored on the days after that date only (the holdout).

\paragraph{Deflated Sharpe ratio}
Let $\widehat{SR}$ be the Sharpe ratio of the simplified DeepSeek-exit rule per lot, over $T$ lots, with sample skewness $\gamma_3$ % num-ok: names (Me2, GT-2), dates, or method parameters, not results
and non-excess kurtosis $\gamma_4$ (both bias-corrected; $\gamma_4=3$ for a normal distribution). With $N$ variants tried and $\sigma^2_{SR}$ the variance of their Sharpe
ratios, the expected maximum Sharpe ratio under the null of no skill is
\[
SR_0=\sigma_{SR}\Bigl((1-\gamma_{\mathrm{EM}})\,\Phi^{-1}\bigl(1-\tfrac1N\bigr)+\gamma_{\mathrm{EM}}\,\Phi^{-1}\bigl(1-\tfrac{1}{Ne}\bigr)\Bigr),
\]
where $\gamma_{\mathrm{EM}}$ is the Euler--Mascheroni constant and $\Phi$ the standard normal distribution function, and the
deflated Sharpe ratio is the probability
\[
\mathrm{DSR}=\Phi\!\left(\frac{(\widehat{SR}-SR_0)\sqrt{T-1}}{\sqrt{1-\gamma_3\widehat{SR}+\frac{\gamma_4-1}{4}
\widehat{SR}^{\,2}}}\right).
\]
At least $N=\dsrNMin$ variants are documented as tried; the true count is not recorded, so we
also report larger $N$. $\sigma^2_{SR}$ is taken from the seven exit variants above, the only set of trials we
can rerun; this is an approximation, since the documented trials also varied entries and the lot cap.

\paragraph{Costs}
Each call is scored as a separate position held for $H$ rows, so it is charged one round trip: the net signed
return is $y-c$ with $c$ equal to \costBp{} basis points (in return units, $c=\costBp/10000$). % num-ok: names (Me2, GT-2), dates, or method parameters, not results

\paragraph{Independent price check}
Ledger prices of NVDA and NDX are compared with daily closes from an independent source, on the same date and on
the previous trading day. The sell-off of October 7 to November 20, 2026 is reported from both. % num-ok: names (Me2, GT-2), dates, or method parameters, not results

\section{Correlated bullish bias}\label{sec:bias}

\paragraph{Sells are rare}
Claude made \nSellClaude{} Sells in \nCallsClaude{} calls (\pctSellClaude\,\%), Gemini \nSellGemini{} in
\nCallsGemini{} (\pctSellGemini\,\%) and DeepSeek \nSellDeepSeek{} in \nCallsDeepSeek{} (\pctSellDeepSeek\,\%)
(Table~\ref{tab:sources}). Claude almost never sells, Gemini rarely does, and DeepSeek is the main seller of the
three.

\paragraph{The three models agree, and err, together}
On the \nThreeWay{} ticker-dates on which all three LLMs called, we compare their calls by ex-ante regime
(Definition~\ref{def:regime}); the error measures use the subset with a known forward return. In calm markets \pctAllBuyCalm\,\% of the three-way calls are unanimous Buys and
\pctAllSellCalm\,\% unanimous Sells; after a fall the shares are \pctAllBuyFall\,\% and \pctAllSellFall\,\%
(Table~\ref{tab:classes}). The mean pairwise correlation of their errors is \rhoErrCalm{} in calm markets,
\rhoErrFall{} after a fall and \rhoErrAll{} overall (Table~\ref{tab:correlation}); mean Cohen's kappa
\cite{cohen1960a} of the signals is \kappaMeanCalm, \kappaMeanFall{} and \kappaMeanAll. In the design-effect sense
of \eqref{eq:neff} the three models are worth $n_{\mathrm{eff}}=\nEffAll$ independent voters.

\begin{figure}[t]
\centering
\includegraphics[width=\columnwidth]{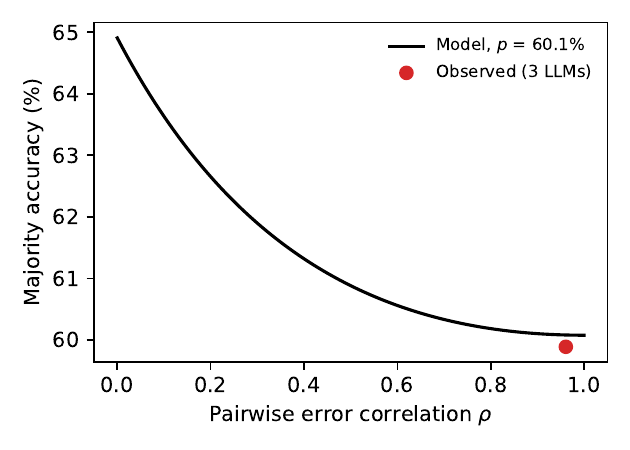}
\caption{Majority accuracy of three voters as a function of their error correlation (Proposition~\ref{prop:jury}),
at the measured mean accuracy, and the observed point for the three LLMs.}
\label{fig:jury}
\end{figure}

\paragraph{What the correlation costs}
Proposition~\ref{prop:jury} turns the measured accuracy and correlation into a majority accuracy. With independent
voters of the same mean accuracy, the majority would be right \accJuryIndepCalm\,\% of the time in calm markets
and \accJuryIndepFall\,\% after a fall. Observed majority accuracy is \accMajCalm\,\% and \accMajFall\,\%. At the
measured correlation the proposition gives almost exactly the mean accuracy of one model, and the observed values
are close to it (Table~\ref{tab:correlation}, Fig.~\ref{fig:jury}). In these data a majority of three LLMs is
about as accurate as one of them.

\paragraph{The bias is bullish}
The shared error has a direction. In calm markets, when the base up-rate is \baseUpCalm\,\%, nearly all Claude and
Gemini calls are Buys, and they are right \accClaudeCalm\,\% and \accGeminiCalm\,\% of the time, below the base
rate (Table~\ref{tab:regimeacc}). After a fall the base up-rate was \baseUpFall\,\%, and the same mostly bullish
calls scored \accClaudeFall\,\% and \accGeminiFall\,\%. Their accuracy moves with the base rate of each regime, below it in calm markets and close to it after a fall.

\begin{figure}[t]
\centering
\includegraphics[width=0.85\columnwidth]{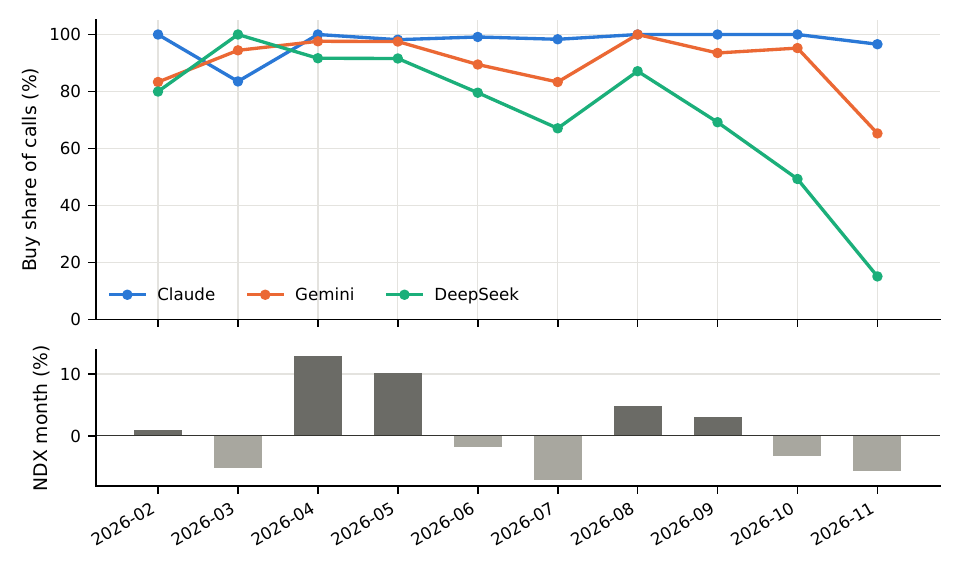}
\caption{Top: share of Buy calls of each LLM by calendar month. Bottom: NDX return of the same month. The last
month is partial (to the end of the sample).}
\label{fig:monthlybuy}
\end{figure}

\begin{figure}[t]
\centering
\includegraphics[width=\columnwidth]{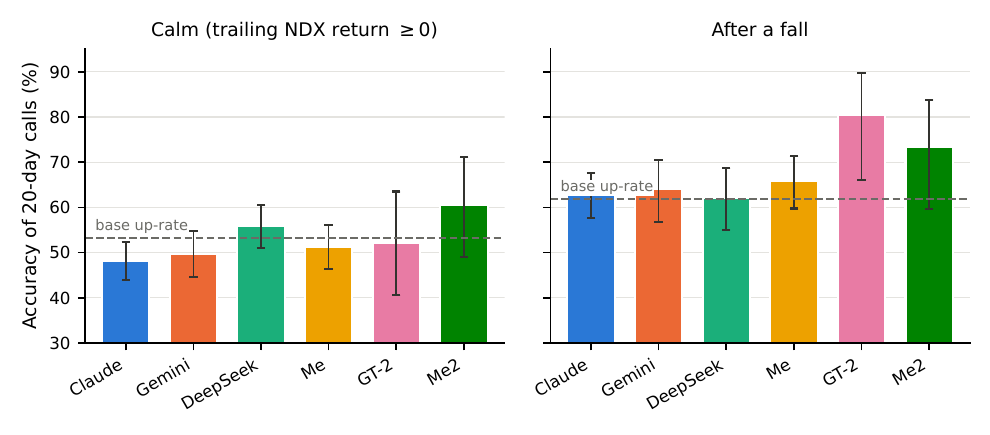}
\caption{Accuracy of each source by ex-ante regime, with 95\,\% Wilson intervals (descriptive; they ignore the % num-ok: confidence level in a caption
overlap of windows). Dashed line: base up-rate of all stock-days in the regime.}
\label{fig:regimeacc}
\end{figure}

\paragraph{The bias over time}
Fig.~\ref{fig:monthlybuy} follows the share of Buy calls of each model month by month. Claude's Buy share is high
in every month, Gemini's is close to it until the last month, and DeepSeek is the only model whose Buy share falls
far during the sample, most clearly in the last months, when NDX fell. Fig.~\ref{fig:regimeacc} shows the regime
pattern of Table~\ref{tab:regimeacc} with intervals. The three LLMs sit near the base up-rate of each regime; the
two combined sources lie above it in places, with wide intervals.

\begin{table*}[t]
\centering
\caption{Agreement and correlated errors of the three LLMs on ticker-dates where all three called, by ex-ante
regime: pairwise error correlations, mean Cohen's $\kappa$, effective number of voters, mean accuracy $\bar p$,
and majority accuracy for independent voters, for correlated voters (Proposition~\ref{prop:jury}) and observed. ``All'' also includes three-way calls in the first
20 trading days, which have no ex-ante regime.} % num-ok: horizon
\label{tab:correlation}
\begin{tabular}{lrrrrrrrrrr}
\toprule
Regime & $n$ & $\rho_{CG}$ & $\rho_{CD}$ & $\rho_{GD}$ & $\bar\kappa$ & $n_\mathrm{eff}$ & $\bar p$ (\%) & Indep.\ (\%) & Corr.\ (\%) & Observed (\%) \\
\midrule
Calm & 100 & 1.00 & 0.96 & 0.96 & 0.66 & 1.02 & 62.0 & 67.7 & 62.0 & 62.0 \\
After a fall & 61 & 1.00 & 0.96 & 0.96 & 0.86 & 1.02 & 66.1 & 73.3 & 66.1 & 65.6 \\
All & 177 & 0.98 & 0.94 & 0.96 & 0.64 & 1.03 & 60.1 & 64.9 & 60.1 & 59.9 \\
\bottomrule
\end{tabular}

\end{table*}

\begin{table}[t]
\centering
\caption{Accuracy and share of Buy calls by ex-ante regime, with the base up-rate of all stock-days.}
\label{tab:regimeacc}
\begin{tabular}{lrrrrrr}
\toprule
 & \multicolumn{3}{c}{Calm} & \multicolumn{3}{c}{After a fall} \\
\cmidrule(lr){2-4}\cmidrule(lr){5-7}
Source & $n$ & Acc.\ (\%) & Buy (\%) & $n$ & Acc.\ (\%) & Buy (\%) \\
\midrule
Claude & 530 & 48.1 & 100 & 349 & 62.8 & 98 \\
Gemini & 358 & 49.7 & 97 & 186 & 64.0 & 88 \\
DeepSeek & 419 & 55.8 & 82 & 190 & 62.1 & 68 \\
Me & 392 & 51.3 & 100 & 257 & 65.8 & 100 \\
GT-2 & 69 & 52.2 & 100 & 41 & 80.5 & 100 \\
Me2 & 71 & 60.6 & 58 & 49 & 73.5 & 53 \\
\midrule
Base up-rate & 2085 & 53.2 & -- & 1313 & 61.9 & -- \\
\bottomrule
\end{tabular}

\end{table}

\section{All-Buy tracks the market}\label{sec:game}

The author's earlier work \cite{zhang2026gametheoretic} sorts the three LLM calls into disagreement classes (all Buy, two Buy and one Sell, and
so on) and treats them as different signals. With the bias of Section~\ref{sec:bias}, the classes collapse. Among three-way calls with a known forward return, the all-Buy class holds
\nAllBuyClsCalm{} calls in calm markets and \nAllBuyClsFall{} after a fall, and no other class holds more than two
(Table~\ref{tab:classes}). The other classes have too few calls to say anything.

The question is then what the all-Buy class is worth. In calm markets it was right \accAllBuyCalm\,\% of the time,
\gapAllBuyCalm{} points above the base up-rate of \baseUpCalm\,\%; after a fall it was right \accAllBuyFall\,\%,
\gapAllBuyFall{} points above \baseUpFall\,\%. These are raw gaps for a call that is long the stock; they carry the
stocks' market exposure, which Section~\ref{sec:me2} estimates per source (the betas of Claude, Gemini and
DeepSeek are all above one, Table~\ref{tab:betaalpha}).

The model of Section~\ref{sec:model} explains why the gap should be small. If each model's call is the sign of a
shared signal plus a common bullish bias plus noise, a strong enough bias makes all-Buy almost certain, and its
accuracy and mean return tend to those of the market itself, whatever the noise correlation
(Proposition~\ref{prop:allbuy}). The model is an illustration: it does not say how large the bias is for these
three models, and the measured gaps are data, not a test of the proposition. What we claim is the measured
statement. The dominant outcome of three LLMs is all-Buy, and its accuracy is close to the base rate of the
market, with a raw residual of a few points.

\begin{table}[t]
\centering
\caption{Accuracy of each disagreement class of the three LLMs and its gap to the directional base rate, by ex-ante regime:
the base up-rate for a Buy majority and the base down-rate (one minus the up-rate) for a Sell majority. Classes with
fewer than five calls are not interpretable.}
\label{tab:classes}
\begin{tabular}{lrrrrrr}
\toprule
 & \multicolumn{3}{c}{Calm} & \multicolumn{3}{c}{After a fall} \\
\cmidrule(lr){2-4}\cmidrule(lr){5-7}
Class & $n$ & Acc.\ (\%) & vs base & $n$ & Acc.\ (\%) & vs base \\
\midrule
All Buy & 97 & 61.9 & +8.6 & 58 & 69.0 & +7.0 \\
2 Buy / 1 Sell & 2 & 50.0 & -3.2 & 1 & 0.0 & -61.9 \\
1 Buy / 2 Sell & 0 & -- & -- & 0 & -- & -- \\
All Sell & 1 & 100.0 & +53.2 & 2 & 0.0 & -38.1 \\
\bottomrule
\end{tabular}

\end{table}

\section{Where \meTwo's edge comes from}\label{sec:me2}

\meTwo{} has the largest share of Sells of all sources (\pctSellMeTwo\,\%, Table~\ref{tab:sources}). After
a fall its Buys were right \accMeTwoBuyFall\,\% of the time ($n=\nMeTwoBuyFall$) and its Sells \accMeTwoSellFall\,\%
($n=\nMeTwoSellFall$); in calm markets the figures are \accMeTwoBuyCalm\,\% and \accMeTwoSellCalm\,\%
(Table~\ref{tab:regimeacc} gives only the combined accuracy). Its Buys did well after a fall, when the base up-rate was \baseUpFall\,\%. Its Sells did not.
They would show foresight of falls, yet they were right less often than its Buys, and with these small samples the
ex-ante regimes give no sign that \meTwo{} times downturns.

\paragraph{Exposure}
Table~\ref{tab:betaalpha} regresses each source's signed 20-day returns on the NDX return \eqref{eq:capm}. % num-ok: horizon
Which source has the lowest per-call beta? \meTwo{} does: its beta is \betaMeTwo, against \betaDeepSeek{} for DeepSeek, \betaMe{} for Me, \betaGemini{} for
Gemini, \betaClaude{} for Claude and \betaGTTwo{} for \gtTwo. Its beta-adjusted alpha is \alphaMeTwo\,\% per 20 days % num-ok: horizon
with a 95\,\% block-bootstrap interval of [\alphaLoMeTwo, \alphaHiMeTwo], which includes zero; the
Newey--West standard error in the same table gives an interval that does not, so the verdict depends on the method. With the bootstrap, only \gtTwo, % num-ok: confidence level
the source \meTwo{} takes its entries from, has an interval above zero (\alphaGTTwo\,\%, [\alphaLoGTTwo,
\alphaHiGTTwo]); with the Newey--West errors, Gemini and DeepSeek do too.

\begin{figure}[t]
\centering
\includegraphics[width=0.7\columnwidth]{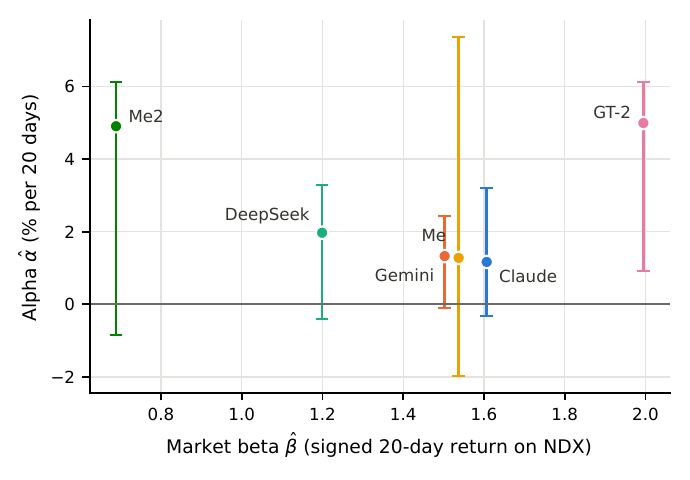}
\caption{Market beta and beta-adjusted alpha per source, with 95\,\% block-bootstrap intervals for alpha % num-ok: confidence level in a caption
(Table~\ref{tab:betaalpha}).}
\label{fig:betaalpha}
\end{figure}

Fig.~\ref{fig:betaalpha} puts the two estimates side by side. The sources line up along the beta axis, and
\meTwo{} sits alone at the low-beta end; its alpha interval is wide and crosses zero. The dollar portfolio below
shows that this low beta is not low exposure of the money invested.

\begin{figure}[t]
\centering
\includegraphics[width=\columnwidth]{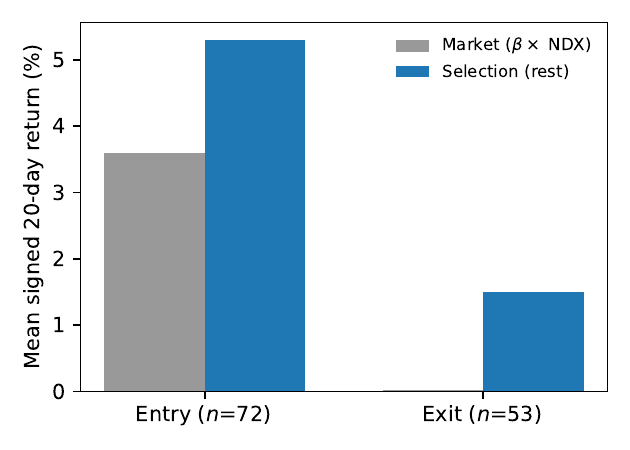}
\caption{\meTwo's mean signed return of entries and exits, split into a market part (each group's own $\hat\beta$ times the market
return) and a selection part. Model-based and descriptive; no interval.}
\label{fig:decomp}
\end{figure}

\paragraph{Entries versus exits}
Fig.~\ref{fig:decomp} splits \meTwo's mean signed return into a market part (the group's own $\hat\beta$ times the market return)
and a selection part. Entries return \meTwoEntryTotal\,\% on average, of which \meTwoEntryMarket{} points are market
and \meTwoEntrySelection{} selection. Exits return \meTwoExitTotal\,\%, of which \meTwoExitMarket{} points are
market and \meTwoExitSelection{} selection. Most of the selection in \meTwo's record comes from its \gtTwo{} entries; its
exits add a smaller part. Each group uses its own beta, since the pooled beta mixes positive-beta Buys and negative-beta Sells; the
split has no interval of its own, so it is descriptive.

\paragraph{Are the exits better than random?}
Of the \nExits{} \meTwo{} exits with a known forward return, \permObs\,\% were followed by a fall over the next
20 rows. Exiting a random stock on the same dates gives \permNull\,\% on average, and the permutation $p$-value is % num-ok: horizon
\permP. The exits are not significantly better than random exits on the same dates; with this sample a modest timing effect cannot be ruled out.

\paragraph{After standard factors}
With the five Fama--French factors and momentum \cite{fama2015a,carhart1997on} (calls whose window ends by
\factorEnd), \meTwo's alpha is \ffAlphaMeTwo\,\% per 20 days, 95\,\% CI [\ffAlphaLoMeTwo, \ffAlphaHiMeTwo] % num-ok: horizon and level
(Table~\ref{tab:factors}). \gtTwo's is \ffAlphaGTTwo\,\% [\ffAlphaLoGTTwo, \ffAlphaHiGTTwo]. Among all sources only
Me, the author's manual calls, keeps an interval above zero (\ffAlphaMe\,\%, [\ffAlphaLoMe, \ffAlphaHiMe]).

\paragraph{The dollar portfolio}
Per-call scoring weights every call equally. Scored instead as the dollar portfolio its trades imply
(Section~\ref{sec:method}, Table~\ref{tab:portfolio}), \meTwo{} looks different. Its alpha against NDX is
\pfAlphaAllMeTwo\,\% a year ($t=\pfTAllMeTwo$, 95\,\% interval [\pfAlphaLoMeTwo, \pfAlphaHiMeTwo]), and the beta % num-ok: confidence level
of its invested dollars is \pfBetaAllMeTwo, not \betaMeTwo. Its low per-call beta is an artifact of scoring: a Sell's signed return is minus the stock's return, so
its \nMeTwoSells{} Sells carry a per-call beta of \betaMeTwoSells{} and cancel most of the \betaMeTwoBuys{} of its
\nMeTwoBuys{} Buys. It is not time in cash either: \meTwo{} held positions on \pfDaysMeTwo{} of \pfDaysAll{}
trading days, as many as \gtTwo{} (\pfDaysGTTwo). When it is invested, it is invested in high-beta stocks. Its portfolio alpha equals that of \gtTwo{}, its entry source (\pfAlphaAllGTTwo\,\% a year,
$t=\pfTAllGTTwo$), so no difference in alpha per invested dollar from the exits is detectable; with intervals this wide, a
modest difference cannot be ruled out. They do limit the capital at risk, and the dollar outcomes
differ: \pfPnlMeTwo{} million dollars for \meTwo{} against \pfPnlGTTwo{} million for \gtTwo, which never sells,
bought more (\pfCostGTTwo{} against \pfCostMeTwo{} million) and also sizes positions differently, so the gap is not
the exits' effect alone. The $t$-statistic of \meTwo's alpha is
at the edge of conventional significance and is not deflated for the six sources examined or the \meTwo{} variants tried.

The edge is concentrated. \meTwo's holding-period P\&L is \meTwoPnl{} million dollars on
\meTwoCost{} million of purchases (\meTwoPnlPct\,\%), of which MU contributes \meTwoPnlMU{} million and AMD
\meTwoPnlAMD{} million; all other tickers together contribute \meTwoPnlRest{} million (Table~\ref{tab:me2pnl}).
Without MU and AMD the portfolio alpha is \pfAlphaAllMeTwoExMuAmd\,\% a year ($t=\pfTAllMeTwoExMuAmd$).

\paragraph{Placebo entries}
Were MU and AMD skill or luck? Table~\ref{tab:placebo} keeps the dates, dollars and sell timing of all
\plNLots{} \meTwo{} lots and swaps only the tickers. When each of the \plNTickers{} real tickers is mapped to a
random one, the placebo portfolios have a median alpha of \plTickAlphaMed\,\% a year and a 95th percentile of
\plTickAlphaHi\,\%; \meTwo's \pfAlphaAllMeTwo\,\% is matched by a share $p=\plTickPAlpha$ of the draws, and its
dollar P\&L by $p=\plTickPPnl$. Over the whole sample, the \gtTwo{} entries picked better stocks than chance. The
holdout does not confirm it yet: on the \plHNLots{} lots bought after 2026-07-26 the alpha is \plHAlphaReal\,\% % num-ok: holdout date
a year against a placebo median of \plHAlphaMed\,\%, with $p=\plHPAlpha$. With so few lots this is a weak test, so
the entry edge is shown in sample and not yet out of sample. Neither $p$-value is deflated for the variants tried.

\paragraph{Where \meTwo{} and \gtTwo{} differ}
Table~\ref{tab:tickerattr} attributes the dollar P\&L of both portfolios by ticker. Both held the two winners:
MU earned \attMeTwoMUPnl{} million dollars for \meTwo{} and \attGTTwoMUPnl{} million for \gtTwo, AMD
\attMeTwoAMDPnl{} and \attGTTwoAMDPnl{} million. The difference is in the losers. \gtTwo, which never sells, bought
\attGTTwoNVDALots{} lots of NVDA, \attGTTwoGOOGLots{} of GOOG and \attGTTwoTSMLots{} of TSM and lost
\attGTTwoNVDAPnl, \attGTTwoGOOGPnl{} and \attGTTwoTSMPnl{} million on them; \meTwo, with its lot cap and exits,
bought \attMeTwoNVDALots, \attMeTwoGOOGLots{} and \attMeTwoTSMLots{} lots and made \attMeTwoNVDAPnl,
\attMeTwoGOOGPnl{} and \attMeTwoTSMPnl{} million. \meTwo{} did not beat \gtTwo{} by picking other stocks; it
bought far fewer lots of the stocks that went wrong; it did not hold them for less time
(Table~\ref{tab:tickerattr}).

\paragraph{Reading}
\meTwo{} does not foresee downturns, and its low per-call beta does not mean low exposure. Its record has two
readings depending on how it is scored. Per call, its alpha is positive, but whether it excludes zero depends on the interval method. As a dollar portfolio,
it has a borderline-significant alpha that comes from its \gtTwo{} entries, mostly two semiconductor positions,
which beat random ticker picks over the whole sample but not yet in the holdout;
its exits add no detectable alpha per dollar but limit how much is invested. The two scorings disagree because they weight differently:
every call equally, or every day's return on the money then invested.

\begin{table*}[t]
\centering
\caption{Per-call and dollar-portfolio scoring of each source. Per call: alpha (\% per 20 days) and beta of signed % num-ok: horizon in a caption
20-day returns (Table~\ref{tab:betaalpha}). Portfolio: daily CAPM on NDX of the return on invested dollars, alpha % num-ok: horizon in a caption
annualised (\% a year), Newey--West $t$ (lag \pfLag) and 95\,\% interval, and holding-period dollar P\&L and
purchases, whole sample.} % num-ok: confidence level in a caption
\label{tab:portfolio}
\begin{tabular}{lrrrrrcrr}
\toprule
 & \multicolumn{2}{c}{Per call (20 d)} & \multicolumn{6}{c}{Dollar portfolio} \\
\cmidrule(lr){2-3}\cmidrule(lr){4-9}
Source & $\alpha$ (\%) & $\beta$ & $\alpha$ (\%/yr) & $t$ & $\beta$ & 95\% CI of $\alpha$ & P\&L (\$M) & Bought (\$M) \\
\midrule
Claude & +1.17 & 1.61 & +16 & +1.09 & 1.51 & [-13, +44] & -6.30 & 54.2 \\
Gemini & +1.33 & 1.50 & +14 & +0.99 & 1.53 & [-14, +42] & -1.83 & 31.2 \\
DeepSeek & +1.97 & 1.20 & +18 & +1.06 & 1.61 & [-15, +51] & -1.85 & 26.9 \\
Me & +1.28 & 1.54 & +14 & +0.85 & 1.30 & [-18, +46] & -2.46 & 41.3 \\
GT-2 & +4.99 & 1.99 & +51 & +2.22 & 1.67 & [+6, +96] & -2.37 & 57.5 \\
Me2 & +4.90 & 0.69 & +51 & +2.00 & 1.63 & [+1, +100] & +4.99 & 36.0 \\
Me2 excl.\ MU, AMD & --- & --- & -8 & -0.48 & 0.87 & [-41, +25] & -0.24 & 30.5 \\
\bottomrule
\end{tabular}

\end{table*}

\begin{table}[t]
\centering
\caption{\meTwo{} holding-period dollar P\&L by ticker, realised and marked at the end of the sample.}
\label{tab:me2pnl}
\begin{tabular}{lr}
\toprule
Ticker & P\&L (\$M) \\
\midrule
MU & +2.74 \\
AMD & +2.50 \\
MRK & +0.31 \\
NOW & +0.24 \\
CRM & +0.07 \\
META & +0.07 \\
TSM & +0.02 \\
AVGO & +0.00 \\
PYPL & -0.05 \\
ORCL & -0.05 \\
AMAT & -0.05 \\
CSCO & -0.06 \\
QCOM & -0.07 \\
ASML & -0.07 \\
MSFT & -0.08 \\
NVDA & -0.22 \\
GOOG & -0.30 \\
\midrule
Total & +4.99 \\
\bottomrule
\end{tabular}

\end{table}

\begin{table}[t]
\centering
\caption{Market beta and beta-adjusted alpha of each source's signed 20-day returns on the NDX 20-day return, with % num-ok: claim labels, horizon, confidence level or dates, not results
date-clustered Newey--West standard errors \cite{newey1987a,driscoll1998consistent} (lag 19) and 20-day % num-ok: claim labels, horizon, confidence level or dates, not results
block-bootstrap \cite{kunsch1989the} 95\,\% intervals.} % num-ok: method parameters in a caption
\label{tab:betaalpha}
\begin{tabular}{lrrrrc}
\toprule
Source & $n$ & $\beta$ & $\alpha$ (\%/20d) & NW s.e. & 95\% block-bootstrap CI \\
\midrule
Claude & 930 & 1.61 & +1.17 & 0.71 & [-0.31, +3.21] \\
Gemini & 603 & 1.50 & +1.33 & 0.54 & [-0.10, +2.43] \\
DeepSeek & 641 & 1.20 & +1.97 & 0.97 & [-0.40, +3.29] \\
Me & 688 & 1.54 & +1.28 & 1.90 & [-1.98, +7.36] \\
GT-2 & 115 & 1.99 & +4.99 & 0.77 & [+0.92, +6.13] \\
Me2 & 125 & 0.69 & +4.90 & 2.22 & [-0.84, +6.13] \\
\bottomrule
\end{tabular}

\end{table}

\begin{table}[t]
\centering
\caption{Alpha after the five Fama--French factors and momentum, calls whose window ends by the last factor date.}
\label{tab:factors}
\begin{tabular}{lrrrcr}
\toprule
Source & $n$ & $\alpha_{\mathrm{FF5+Mom}}$ (\%/20d) & NW s.e. & 95\% CI & $\beta_{\mathrm{Mkt}}$ \\
\midrule
Claude & 606 & +1.23 & 0.81 & [-0.35, +2.81] & 2.09 \\
Gemini & 430 & +0.88 & 0.92 & [-0.92, +2.68] & 2.21 \\
DeepSeek & 485 & +0.50 & 0.68 & [-0.82, +1.83] & 2.79 \\
Me & 526 & +3.96 & 1.33 & [+1.35, +6.57] & 1.87 \\
GT-2 & 92 & +2.62 & 2.20 & [-1.70, +6.94] & 3.43 \\
Me2 & 87 & -0.67 & 1.92 & [-4.44, +3.09] & 3.65 \\
\bottomrule
\end{tabular}

\end{table}

\begin{table}[t]
\centering
\caption{Placebo entries: \meTwo's lots with their dates, dollars and sell timing kept and their tickers drawn at
random (\plDraws{} draws). One-sided $p$: share of draws at least as good as \meTwo.}
\label{tab:placebo}
\begin{tabular}{lrrrrrr}
\toprule
 & \multicolumn{3}{c}{Portfolio $\alpha$ (\%/yr)} & \multicolumn{3}{c}{Dollar P\&L (\$M)} \\
\cmidrule(lr){2-4}\cmidrule(lr){5-7}
Placebo & Median & 95th pct. & $p$ & Median & 95th pct. & $p$ \\
\midrule
Random ticker per lot & +13 & +32 & 0.001 & +1.33 & +3.01 & 0.001 \\
Random ticker per real ticker & +12 & +40 & 0.010 & +1.25 & +3.84 & 0.009 \\
\midrule
\meTwo{} (actual) & \multicolumn{3}{c}{+51} & \multicolumn{3}{c}{+4.99} \\
\bottomrule
\end{tabular}

\end{table}

\begin{table*}[t]
\centering
\caption{Dollar P\&L by ticker for \meTwo{} and \gtTwo: lots bought, dollars bought, dollar-weighted mean trading
days held per lot (open lots to the end of the sample) and P\&L, sorted by the difference in P\&L.}
\label{tab:tickerattr}
\begin{tabular}{lrrrrrrrr}
\toprule
 & \multicolumn{4}{c}{\meTwo} & \multicolumn{4}{c}{\gtTwo} \\
\cmidrule(lr){2-5}\cmidrule(lr){6-9}
Ticker & Lots & Bought (\$M) & Days & P\&L (\$M) & Lots & Bought (\$M) & Days & P\&L (\$M) \\
\midrule
NVDA & 5 & 2.5 & 122 & -0.22 & 19 & 9.5 & 107 & -2.02 \\
GOOG & 8 & 4.0 & 54 & -0.30 & 15 & 7.5 & 116 & -1.58 \\
TSM & 5 & 2.5 & 118 & +0.02 & 14 & 7.0 & 98 & -0.90 \\
MU & 6 & 3.0 & 112 & +2.74 & 11 & 5.5 & 139 & +2.11 \\
AVGO & 5 & 2.5 & 71 & +0.00 & 7 & 3.5 & 110 & -0.55 \\
AMD & 5 & 2.5 & 149 & +2.50 & 10 & 5.0 & 120 & +1.97 \\
NOW & 4 & 2.0 & 15 & +0.24 & 4 & 2.0 & 92 & -0.27 \\
QCOM & 5 & 2.5 & 11 & -0.07 & 5 & 2.5 & 122 & -0.49 \\
ORCL & 3 & 1.5 & 7 & -0.05 & 3 & 1.5 & 52 & -0.35 \\
CRM & 6 & 3.0 & 44 & +0.07 & 6 & 3.0 & 69 & -0.16 \\
META & 4 & 2.0 & 19 & +0.07 & 4 & 2.0 & 76 & -0.07 \\
ASML & 2 & 1.0 & 17 & -0.07 & 2 & 1.0 & 42 & -0.21 \\
AMAT & 1 & 0.5 & 5 & -0.05 & 1 & 0.5 & 52 & -0.13 \\
MRK & 5 & 2.5 & 82 & +0.31 & 6 & 3.0 & 96 & +0.25 \\
CSCO & 1 & 0.5 & 44 & -0.06 & 1 & 0.5 & 44 & -0.06 \\
PYPL & 2 & 1.0 & 34 & -0.05 & 2 & 1.0 & 84 & -0.03 \\
MSFT & 5 & 2.5 & 18 & -0.08 & 5 & 2.5 & 61 & +0.12 \\
\midrule
Total & 72 & 36.0 & & +4.99 & 115 & 57.5 & & -2.37 \\
\bottomrule
\end{tabular}

\end{table*}

\section{The dissenting source and exposure}\label{sec:dissent}

This section is preliminary: it rests on one dissenting source and one bear episode.

DeepSeek is the main seller among the three LLMs (\pctSellDeepSeek\,\% of its calls), and \meTwo{} uses its Sells
only as exits. To isolate that choice we run a simplified \meTwo{} rule, which sells all lots of a ticker on an exit, on the ledger signals and change only the
exit (Section~\ref{sec:me2policy}, Table~\ref{tab:counterfactual}). The replay covers the ledger: all
\cfLedgerBuys{} \meTwo{} Buys in the ledger are among its \cfReplayBuys{} entries; the extra entries come from
replaying on ledger signals only.

\paragraph{Which exit works}
With the DeepSeek Sell as exit, the \cfNDs{} lots returned \cfMeanDs\,\% on average over a mean
holding of \cfHoldDs{} ledger rows, with beta \cfBetaDs{} and alpha \cfAlphaDs\,\% (95\,\% CI [\cfAlphaLoDs, % num-ok: confidence level
\cfAlphaHiDs]). Without any exit, lots returned \cfMeanNone\,\% and were held \cfHoldNone{} rows on average. With the
Claude or the Gemini Sell as exit, lots returned \cfMeanCl\,\% and \cfMeanGe\,\%; many of their lots were still open at the
end (Table~\ref{tab:counterfactual}), because those models seldom sell, and their alpha intervals are very wide
and include zero. Exiting on a Sell from any of the three LLMs returned \cfMeanAny\,\%, close to the DeepSeek exit,
since most LLM Sells are DeepSeek's (\nSellDeepSeek{} of its calls, against \nSellClaude{} for Claude and
\nSellGemini{} for Gemini).

\begin{figure}[t]
\centering
\includegraphics[width=\columnwidth]{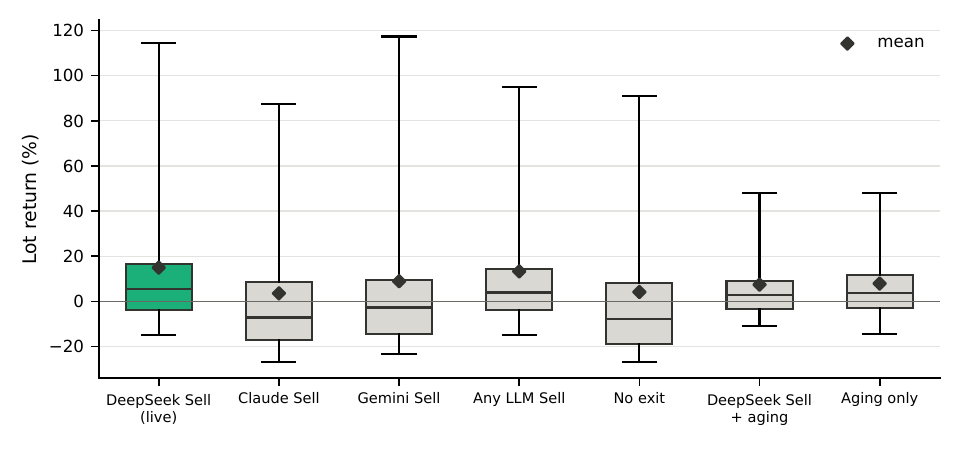}
\caption{Lot returns of the simplified \meTwo{} rule replayed with each exit rule: quartile box, whiskers at the 5th % num-ok: percentile definition, not a result
and 95th percentiles, diamond at the mean. Highlighted: the DeepSeek exit.} % num-ok: percentile definition, not a result
\label{fig:exitreturns}
\end{figure}

Fig.~\ref{fig:exitreturns} shows the whole distribution of lot returns behind these means. The DeepSeek exit
moves the middle of the distribution up, with a median above zero. The Claude, Gemini and no-exit variants have
medians below zero and long right tails from lots that were never sold, which is why their means are fragile.

\paragraph{What the exit does not do}
The DeepSeek exit does not lower the beta of a lot; it is slightly higher, \cfBetaDs{} with the exit
against \cfBetaNone{} without. Lot
betas are high for every exit rule because a lot is long one stock for weeks, so this beta is not comparable with
the per-call betas of Table~\ref{tab:betaalpha}. The aging exit, which sold every lot after a fixed number of
sessions and was removed on 26 July, gave a lower lot beta (\cfBetaAgeonly) and lower mean returns
(\cfMeanAgeonly\,\%).

\paragraph{Reading}
Among exits for the same entry rule, the exits that close lots often raised the win rate and the return per row
held: the DeepSeek Sell (\cfPerRowDs\,\% per row), any LLM Sell (\cfPerRowAny\,\%) and the aging exit
(\cfPerRowAgeonly\,\%), against \cfPerRowNone\,\% with no exit (Table~\ref{tab:counterfactual}). The Claude and
Gemini Sells, which are rare, did not. Among the model Sells, only DeepSeek's helps.
The exit raised returns without lowering exposure. With one dissenting source,
the exit's effect cannot be separated from that source's particular behaviour in this sample, and the exit variants
tried here enter the deflated Sharpe ratio of Section~\ref{sec:live}.

\begin{table*}[t]
\centering
\caption{A simplified \meTwo{} rule (all lots sold on an exit) run on the ledger signals with only the exit changed. Lots: lots bought; Open: lots still open at
the end of the sample (marked at the last price); Hold: mean holding in ledger rows; Win: share of lots with a positive
return; $\beta$ and $\alpha$ of lot returns on the NDX return over the same days, with 95\,\% block-bootstrap % num-ok: confidence level in a caption
intervals.}
\label{tab:counterfactual}
\begin{tabular}{lrrrrrrrrc}
\toprule
Exit rule & Lots & Open & Hold (d) & Win (\%) & Mean (\%) & Per row (\%) & $\beta$ & $\alpha$ (\%) & 95\% CI \\
\midrule
DeepSeek Sell, all lots & 86 & 1 & 38.2 & 65.1 & +14.93 & +0.39 & 3.21 & +5.50 & [+1.02, +7.16] \\
Claude Sell & 70 & 57 & 107.0 & 32.9 & +3.56 & +0.03 & 2.86 & +16.57 & [-17.38, +20.71] \\
Gemini Sell & 71 & 30 & 80.5 & 43.7 & +8.94 & +0.11 & 2.40 & +12.00 & [-2.58, +15.40] \\
Any LLM Sell & 88 & 0 & 35.6 & 63.6 & +13.28 & +0.37 & 3.30 & +5.63 & [+1.87, +7.35] \\
No exit (hold to end) & 67 & 67 & 115.5 & 32.8 & +4.18 & +0.04 & 2.94 & +17.34 & [-15.07, +23.79] \\
DeepSeek Sell + 20-session aging & 112 & 0 & 15.9 & 61.6 & +7.40 & +0.47 & 2.24 & +3.63 & [+0.79, +4.84] \\
20-session aging only & 112 & 0 & 20.0 & 62.5 & +7.91 & +0.40 & 2.00 & +4.73 & [+0.40, +6.22] \\
\bottomrule
\end{tabular}

\end{table*}

\section{Robustness}\label{sec:live}

This section uses the whole sample. The \meTwo{} rules were developed on the same ledger, so every \meTwo{} result is
in sample; this section reports how sensitive the results are to costs, to the number of variants tried, to the
entry price and to the ledger prices.

\paragraph{Costs}
With a round-trip cost of \costBp{} basis points per call, \meTwo's mean signed return per call falls from
\grossMeTwoAll\,\% to \netMeTwoAll\,\%, and \gtTwo's from \grossGTTwoAll\,\% to \netGTTwoAll\,\%
(Table~\ref{tab:costs}). Costs at this level do not change any sign.

\paragraph{Multiple testing}
At least \dsrNMin{} variants of \meTwo{} were tried (Appendix~\ref{app:variants}); the true number is not
recorded. How much does that matter? The per-lot Sharpe ratio of the simplified DeepSeek-exit rule (Table~\ref{tab:counterfactual}) is \dsrSharpe{} over \dsrLots{}
lots. Its deflated Sharpe ratio \cite{bailey2014the} is \dsrMin{} if \dsrNMin{} variants were tried, \dsrTwenty{}
for twenty and \dsrFifty{} for fifty (Table~\ref{tab:deflated}). The lot-level result survives deflation for up to
twenty variants and is borderline at fifty. It leans on the strong positive skew of lot returns, and the lots
overlap in time and ticker, so the effective number of lots is smaller than $T$. The per-call alpha of Section~\ref{sec:me2} excludes zero only under the
Newey--West interval, and is not deflated either. The dollar-portfolio alpha ($t=\pfTAllMeTwo$) is not deflated; it barely
clears the conventional level, so any correction for the variants above would put it below.

\begin{figure}[t]
\centering
\includegraphics[width=0.7\columnwidth]{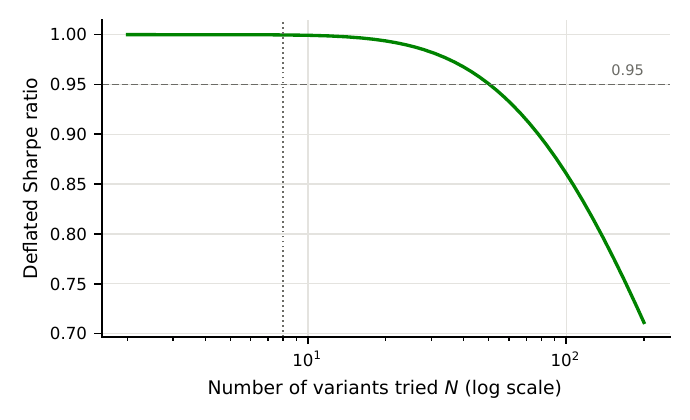}
\caption{Deflated Sharpe ratio of the simplified \meTwo{} DeepSeek-exit rule per lot as a function of the number of variants tried
(log scale). Dotted line: documented minimum number of variants; dashed line: 0.95.} % num-ok: conventional threshold
\label{fig:dsr}
\end{figure}

Fig.~\ref{fig:dsr} traces the deflated Sharpe ratio over the number of variants: it is \dsrTwenty{} at
twenty variants and is just above the conventional level of 0.95 at fifty (\dsrFifty), a borderline pass. % num-ok: conventional threshold

\paragraph{Entry price}
Because the ledger price is the previous close, a trader acting on a call could only get the close of the signal
day. Entering one ledger row later changes little (Table~\ref{tab:entrydelay}): \meTwo's beta becomes
\betaMeTwoNext{} and its alpha \alphaMeTwoNext\,\% with interval [\alphaLoMeTwoNext, \alphaHiMeTwoNext], still
including zero; \gtTwo's alpha becomes \alphaGTTwoNext\,\% with interval [\alphaLoGTTwoNext, \alphaHiGTTwoNext].

\paragraph{Independent prices}
Table~\ref{tab:pricecheck} compares the ledger with independent daily closes for NVDA and NDX. The median gap is
zero once the ledger is shifted by one trading day, and the shift is exact on \exactLagNvda\,\% of days. On
independent closes the sell-off from 7 October to 20 November was \trueSelloffNvda\,\% for NVDA and % num-ok: sell-off dates
\trueSelloffNdx\,\% for NDX.

\begin{table}[t]
\centering
\caption{Scored calls, accuracy, and mean signed 20-day return per call over the whole sample, gross and net of a % num-ok: horizon in a caption
fixed round-trip cost (\%).}
\label{tab:costs}
\begin{tabular}{lrrrr}
\toprule
Source & $n$ & Acc.\ (\%) & Gross & Net \\
\midrule
Claude & 930 & 54.0 & +1.83 & +1.63 \\
Gemini & 603 & 53.6 & +3.15 & +2.95 \\
DeepSeek & 641 & 56.0 & +3.15 & +2.95 \\
Me & 688 & 56.2 & +2.59 & +2.39 \\
GT-2 & 115 & 61.7 & +8.06 & +7.86 \\
Me2 & 125 & 64.8 & +5.76 & +5.56 \\
\bottomrule
\end{tabular}

\end{table}

\begin{table}[t]
\centering
\caption{Deflated Sharpe ratio of the simplified \meTwo{} DeepSeek-exit rule per lot, for several numbers of variants tried.}
\label{tab:deflated}
\begin{tabular}{rrrr}
\toprule
Variants $N$ & Sharpe per lot & Expected max.\ Sharpe & Deflated Sharpe (prob.) \\
\midrule
8 & 0.42 & 0.21 & 1.000 \\
20 & 0.42 & 0.27 & 0.994 \\
50 & 0.42 & 0.32 & 0.951 \\
\bottomrule
\end{tabular}

\end{table}

\begin{table*}[t]
\centering
\caption{Main estimates with entry at the prior close (main convention) and at the close of the signal day (one
ledger row later).}
\label{tab:entrydelay}
\begin{tabular}{lrrrrll}
\toprule
 & \multicolumn{2}{c}{Accuracy (\%)} & \multicolumn{2}{c}{$\beta$} & \multicolumn{2}{c}{$\alpha$ (\%/20d)} \\
\cmidrule(lr){2-3}\cmidrule(lr){4-5}\cmidrule(lr){6-7}
Source & Prior close & Signal-day close & Prior & Signal-day & Prior [95\% CI] & Signal-day [95\% CI] \\
\midrule
Claude & 54.0 & 53.0 & 1.61 & 1.58 & +1.17 [-0.31, +3.21] & +0.99 [-0.44, +2.94] \\
Gemini & 53.6 & 53.3 & 1.50 & 1.53 & +1.33 [-0.10, +2.43] & +1.03 [-0.22, +2.09] \\
DeepSeek & 56.0 & 54.6 & 1.20 & 1.17 & +1.97 [-0.40, +3.29] & +2.02 [-0.29, +3.45] \\
GT-2 & 61.7 & 64.3 & 1.99 & 2.05 & +4.99 [+0.92, +6.13] & +5.62 [+2.31, +6.55] \\
Me2 & 64.8 & 66.9 & 0.69 & 0.84 & +4.90 [-0.84, +6.13] & +4.51 [-0.17, +6.10] \\
\bottomrule
\end{tabular}

\end{table*}

\begin{table}[t]
\centering
\caption{Ledger prices against independent daily closes: median absolute gap on the same day and on the previous
trading day, share of exact matches after the shift, and the sell-off on both.}
\label{tab:pricecheck}
\begin{tabular}{lrrrrrr}
\toprule
 & & \multicolumn{2}{c}{Median $|$ledger/independent$-1|$ (\%)} & Exact after & \multicolumn{2}{c}{Oct 7 $\to$ Nov 20 (\%)} \\
\cmidrule(lr){3-4}\cmidrule(lr){6-7}
Series & Days & Same day & Previous day & shift (\%) & Ledger & Independent \\
\midrule
NVDA & 182 & 1.51 & 0.00 & 86 & -39.0 & -47.1 \\
NDX & 180 & 0.74 & 0.00 & 86 & -15.1 & -19.7 \\
\bottomrule
\end{tabular}

\end{table}

\section{What transfers to production}\label{sec:production}

Which parts of the system would be worth keeping in a production setting? We test three candidates. Each is scored
on the whole sample and, separately, on the holdout: the days after 26 July 2026, when the current \meTwo{} rules % num-ok: holdout date
took effect; from then on the records were produced as the system ran. Nothing in this section was tuned on the
holdout. It is short and contains one sell-off, so a holdout
result is a check of direction, not a proof.

\paragraph{Position limits}
For each of the \prodNSrc{} signal sources, a portfolio is simulated on the ledger signals with lots of
\prodLotK{} thousand dollars at the ledger price, under three policies: buy a lot on every Buy and never sell
(raw); at most \prodCap{} open lots per ticker (cap); and the cap plus selling the oldest lot on a DeepSeek Sell
(cap and exit), which is \meTwo's rule applied to another entry source. Applied to \gtTwo, the raw policy reproduces
\gtTwo's ledger portfolio exactly, and the cap and exit come close to \meTwo's ledger record. All
\prodNVariants{} variants are reported (Table~\ref{tab:prodlimits}).

The cap and exit improve the dollar result for all \prodNSrc{} sources over the whole sample, also per dollar
bought (\prodCapExitRetWins{} of \prodNSrc), and lose less in the holdout for all \prodCapExitWinsH. For Claude, buying every Buy lost \prodPnlClaudeRaw{} million dollars on
\prodCostClaudeRaw{} million of purchases; with the cap and exit, Claude's signals made \prodPnlClaudeCapExit{}
million on \prodCostClaudeCapExit{} million. In the holdout the same comparison is \prodPnlHClaudeRaw{} against
\prodPnlHClaudeCapExit{} million. The limits do not raise alpha, however: Claude's falls from
\prodAlphaAllClaudeRaw\,\% to \prodAlphaAllClaudeCapExit\,\% a year, and \gtTwo's stays near
\prodAlphaAllGTTwoCapExit\,\%. They work by keeping less money in positions that keep falling. In the holdout, which ends in a sell-off, all
\prodNLossH{} variants lose money and \prodNNegAlphaH{} of \prodNVariants{} have a negative alpha; \gtTwo's
alpha there is \prodAlphaHGTTwoRaw\,\%, \prodAlphaHGTTwoCap\,\% and \prodAlphaHGTTwoCapExit\,\% a year
under the three policies, so its whole-sample alpha does not carry over. In the holdout the cap alone has a
higher alpha than the cap and exit for \prodCapBeatsExitH{} of \prodNSrc{} sources. Position limits are
therefore risk control that loses less, not an edge of their own.

\paragraph{Stock picking out of sample}
Section~\ref{sec:me2} found that the \gtTwo{} entries picked better stocks than chance over the whole sample. On the
\plHNLots{} \meTwo{} lots bought in the holdout, the placebo gives $p=\plHPAlpha$. The stock-picking edge is not
confirmed out of sample; with so few lots the test has little power, so it is also not refuted. These lots
earn an alpha of \plHAlphaReal\,\% a year but lose \plHPnlReal{} million dollars; the \gtTwo{} cap-and-exit
portfolio in Table~\ref{tab:prodlimits} has a holdout alpha of \prodAlphaHGTTwoCapExit\,\% because it also
carries positions bought before the holdout.

\paragraph{DeepSeek's Sells as a market signal}
DeepSeek is the only one of the three models whose share of Sells moves (Fig.~\ref{fig:monthlybuy}). On each date
we regress the NDX return over the next 20 rows on the share of DeepSeek's calls that are Sells, with Newey--West % num-ok: horizon
errors for the overlapping windows (Table~\ref{tab:prodsellshare}). More DeepSeek Sells are followed by lower NDX
returns: the slope has $t=\ssTDeepSeekAll$ over the whole sample (correlation \ssCorrDeepSeekAll) and keeps its
sign in the holdout ($t=\ssTDeepSeekH$). With the trailing 20-row NDX return as a control, so that the Sell % num-ok: horizon
share is not just a sign of a recent fall, the slope stays negative ($t=\ssTCtlDeepSeekAll$ whole sample,
\ssTCtlDeepSeekH{} holdout). This was the one market signal we expected before running the test. It rests on
a median of \ssMedCallsDeepSeekAll{} DeepSeek calls per date. The other models behave differently: Claude's
share of Sells has the opposite sign in the holdout ($t=\ssTClaudeH$), but with the same control it falls to
\ssTCtlClaudeH, and Claude sells so rarely that a few dates drive it; Gemini's is weak. With six regressions and no deflation, the
DeepSeek result is borderline. If it holds on more data, its use would be to set how much the whole system
invests, not which stocks it buys.

\paragraph{Reading}
Of the three candidates, position limits transfer cleanly, as risk control; DeepSeek's share of Sells is a
promising but borderline market signal; and the stock-picking edge of \gtTwo{} is shown only in sample. The
practical value of the data is less in any one model's picks than in what the models' disagreement says about the
market, and in the discipline of limiting positions.

\begin{table*}[t]
\centering
\caption{Position limits for each entry source: buy on every Buy and never sell (raw), at most \prodCap{} open lots
per ticker (cap), and the cap plus selling the oldest lot on a DeepSeek Sell (cap + exit). Daily alpha against NDX
(\% a year, Newey--West $t$), dollar P\&L and dollars bought, on the whole sample and on the holdout after the
current \meTwo{} rules took effect.}
\label{tab:prodlimits}
\begin{tabular}{llrrrrrrrr}
\toprule
 & & \multicolumn{5}{c}{Whole sample} & \multicolumn{3}{c}{Holdout} \\
\cmidrule(lr){3-7}\cmidrule(lr){8-10}
Entry source & Policy & $\alpha$ (\%/yr) & $t$ & P\&L (\$M) & Bought (\$M) & P\&L/bought (\%) & $\alpha$ (\%/yr) & $t$ & P\&L (\$M) \\
\midrule
Claude & raw & +14 & +0.95 & -63.86 & 544.2 & -11.7 & -22 & -1.15 & -97.93 \\
Claude & cap & +9 & +0.62 & -0.43 & 54.5 & -0.8 & -14 & -0.72 & -9.34 \\
Claude & cap + exit & +8 & +0.54 & +1.47 & 114.4 & +1.3 & -25 & -1.20 & -7.96 \\
\addlinespace
Gemini & raw & +13 & +0.91 & -22.88 & 313.4 & -7.3 & -19 & -1.13 & -55.29 \\
Gemini & cap & +12 & +0.80 & -0.49 & 56.0 & -0.9 & -10 & -0.54 & -8.58 \\
Gemini & cap + exit & +12 & +0.77 & +2.83 & 103.0 & +2.7 & -23 & -1.16 & -5.91 \\
\addlinespace
DeepSeek & raw & +17 & +1.03 & -30.84 & 270.4 & -11.4 & -26 & -1.26 & -52.22 \\
DeepSeek & cap & +17 & +1.12 & -0.19 & 53.0 & -0.4 & -13 & -0.64 & -8.67 \\
DeepSeek & cap + exit & +14 & +0.84 & +3.78 & 95.0 & +4.0 & -31 & -1.40 & -5.39 \\
\addlinespace
Me & raw & +14 & +0.86 & -24.92 & 414.8 & -6.0 & +10 & +0.40 & -39.06 \\
Me & cap & +6 & +0.31 & -1.73 & 41.0 & -4.2 & -16 & -0.62 & -6.89 \\
Me & cap + exit & +11 & +0.65 & +1.85 & 107.5 & +1.7 & -10 & -0.39 & -3.96 \\
\addlinespace
GT-2 & raw & +51 & +2.22 & -2.37 & 57.5 & -4.1 & -16 & -0.82 & -9.93 \\
GT-2 & cap & +52 & +2.30 & +1.40 & 33.5 & +4.2 & -11 & -0.52 & -5.22 \\
GT-2 & cap + exit & +54 & +2.19 & +5.75 & 38.0 & +15.1 & -25 & -0.87 & -1.40 \\
\bottomrule
\end{tabular}

\end{table*}

\begin{table}[t]
\centering
\caption{Share of each LLM's calls that are Sells, per date, against the NDX return over the next 20 rows: OLS % num-ok: horizon in a caption
slope (percentage points of NDX return per 100\,\% Sells), Newey--West $t$ (lag 19), the $t$ of the same slope with
the trailing 20-row NDX return as a control, median calls per date, and correlation.} % num-ok: method parameters in a caption
\label{tab:prodsellshare}
\begin{tabular}{llrrrrrr}
\toprule
LLM & Period & Dates & Calls/date & Slope (pp per 100\% Sells) & NW $t$ & $t$, trailing-NDX control & Corr. \\
\midrule
DeepSeek & whole & 164 & 4 & -8.7 & -2.11 & -2.26 & -0.35 \\
DeepSeek & holdout & 61 & 4 & -4.8 & -1.50 & -2.77 & -0.26 \\
Claude & whole & 160 & 5 & +7.9 & +1.88 & +2.20 & +0.13 \\
Claude & holdout & 62 & 8 & +41.0 & +3.44 & +0.77 & +0.18 \\
Gemini & whole & 160 & 4 & +1.8 & +0.45 & +0.73 & +0.04 \\
Gemini & holdout & 59 & 4 & +11.4 & +1.95 & +1.07 & +0.19 \\
\bottomrule
\end{tabular}

\end{table}

\section{Discussion and limitations}\label{sec:discussion}

\paragraph{What the results show}
On the ticker-dates where all three called, three LLMs from different providers behaved almost like one voter
with a bullish lean ($n_{\mathrm{eff}}=\nEffAll$). Their agreement is not evidence of correctness, and their
majority was about as accurate as their mean member (Table~\ref{tab:correlation}). A policy that takes Buys from
one combination and Sells from the one model that dissents looks low-beta per call only because its Sells enter
with negative beta; the dollars it invests carry a beta of \pfBetaAllMeTwo. Its exits show no timing skill ($p=\permP$). Its dollar-portfolio alpha is
borderline significant and comes from its entries, mostly two positions; its exits add no detectable alpha per dollar but
limit the capital at risk, and \meTwo{} ended with a dollar gain where its entry source lost; the two
portfolios also differ in position sizing, so this difference is not the exits' effect alone. Finally, per-call and dollar-portfolio scoring gave
different verdicts on the same policy; evaluations of LLM trading signals should report both.

\paragraph{Limitations}
\begin{itemize}
\item \emph{Small samples.} \meTwo{} has \nCallsMeTwo{} calls and \nExits{} scored exits; \gtTwo{} has \nCallsGTTwo{} calls. Intervals for these two sources are wide.
\item \emph{One bear episode.} The sample contains one sell-off. Whether the patterns hold in another downturn,
and out of sample for the \meTwo{} rules, is untested.
\item \emph{One dissenting source.} The exit result of Section~\ref{sec:dissent} rests on DeepSeek alone and is
preliminary.
\item \emph{Unknown number of variants.} The deflated Sharpe ratio is reported for several counts because the
true number of \meTwo{} variants tried is not recorded, and the variance of trial Sharpe ratios is approximated
from the exit variants only.
\item \emph{Survivorship.} The ticker list holds large stocks that exist today, chosen and extended by the
author during the sample.
Results may be more favourable than on a survivorship-free universe.
\item \emph{Ledger prices.} The ledger records the previous close and a few non-trading rows; we correct for the
first in every return and check robustness to the entry price, but the ledger is the only source for most tickers.
\item \emph{Factor window.} The factor data end on \factorEnd, before the October--November sell-off, so the factor-adjusted
alphas do not cover it.
\item \emph{Daily weighting.} The dollar-portfolio regression weights every day equally, however much capital
is invested that day, so a source can show a positive alpha and still lose money in dollars (Table~\ref{tab:portfolio}).
\item \emph{Holdout.} The holdout of Section~\ref{sec:production} is a few months with one sell-off, and the six
sell-share regressions are not deflated; its results are checks of direction.
\item \emph{Concentration.} \meTwo's dollar edge rests on two tickers, MU and AMD; without them its portfolio
alpha is \pfAlphaAllMeTwoExMuAmd\,\% a year. A result carried by two positions is weak evidence for a rule.
\item \emph{Partly negative result.} The contribution on \meTwo{} is an explanation, not a profitable strategy:
whether its per-call alpha excludes zero depends on the interval method, its portfolio alpha is borderline and undeflated, and all results are in
sample, because the rules were developed on the same data. We report it as such, in the spirit
of the author's earlier negative-results study of an equity screen \cite{zhang2026failure}.
\end{itemize}

\section{Conclusion}\label{sec:conclusion}

Combining several LLMs assumes they err independently. On live trading calls they do not: their errors are highly
correlated ($\rho=\rhoErrAll$), they lean bullish together, and their majority is about as accurate as their
mean member.
The dominant outcome, all three saying Buy, is right a few points more often than the market rises (\gapAllBuyCalm{} and \gapAllBuyFall{} points, raw). A role-split policy
that exits on DeepSeek's Sells has the lowest per-call beta of all sources, yet its invested dollars carry a beta of
\pfBetaAllMeTwo: its low per-call beta comes from scoring its Sells as negative-beta calls, not from defensive holdings. Per call, whether its alpha excludes zero depends on
the interval method; as a dollar portfolio it earns \pfAlphaAllMeTwo\,\% a year over NDX ($t=\pfTAllMeTwo$, undeflated), matching
its game-theoretic entry source and coming mostly from two positions; random ticker picks with the same dates and dollars rarely match it
($p=\plTickPAlpha$), though the holdout does not yet confirm it. Its exits add no detectable alpha per
invested dollar but
limit the capital at risk; its dollar P\&L is \pfPnlMeTwo{} million against \pfPnlGTTwo{} million for its entry source,
which also sizes positions differently. These results are in sample: the rules were developed on the same data. For production, the parts that transfer are position limits, which helped every
signal source as risk control, and possibly DeepSeek's share of Sells as a market signal; the stock-picking edge
is not yet confirmed out of sample. For multi-LLM trading systems the practical lessons are to measure the
correlation and direction of the models' errors before counting their votes, and to score a policy both per call
and as the dollar portfolio it actually holds, since the two can disagree.

\section*{Data availability}
The LLM and manual calls analysed here were recorded by the author's Manual\_Stock system, and the \gtTwo{} and
\meTwo{} records were computed from them by their rules; the system's web dashboard is
at \url{https://manual-stock.duckdns.org/web/index.html}; access requires a login. The ledger snapshot used in the
paper (to the end of the sample), the analysis scripts that rebuild every table, figure and number, and the
independent price and factor files with their checksums are kept with the paper's source.

\appendices

\section{Variants of \meTwo{} tried}\label{app:variants}

The variants documented are: the stress filter on and off; the unanimous-LLM entry rule; the
aging exit on and off; and caps of two, four and five open lots per ticker, and no cap. These give at least
\dsrNMin{} configurations. The exit variants of Table~\ref{tab:counterfactual} were run for this paper and are used
for the variance of trial Sharpe ratios in the deflated Sharpe ratio.

\section{Notation}\label{app:notation}

\begin{table}[h]
\centering
\caption{Main notation.}
\label{tab:notation}
\begin{tabular}{ll}
\toprule
Symbol & Meaning \\
\midrule
$c=(d,i,s,g)$ & a call: date, ticker, source, signal ($g=+1$ Buy, $-1$ Sell) \\
$H$ & horizon in ledger rows \\
$R_i(d)$, $m(d)$ & forward return of ticker $i$ and of NDX over $H$ rows from $d$ \\
$y=g\,R_i(d)$ & signed return of a call \\
$p$, $\rho$ & mean accuracy and mean pairwise error correlation of the three LLMs \\
$n_{\mathrm{eff}}$ & effective number of independent voters, $3/(1+2\rho)$ \\
$M(\rho)$ & majority accuracy of three correlated voters (Proposition~\ref{prop:jury}) \\
$b$, $z$, $e_j$ & common bullish bias, shared signal and noise of model $j$ (Section~\ref{sec:model}) \\
$\alpha$, $\beta$ & intercept and market slope of signed returns on the market return \\
$\widehat{SR}$, $SR_0$, DSR & observed Sharpe ratio, expected maximum under the null, deflated Sharpe ratio \\
$N$ & number of policy variants tried \\
\bottomrule
\end{tabular}
\end{table}
\FloatBarrier

\section*{Acknowledgments}
\emph{Author contributions.} The author designed and built the Manual\_Stock system, has collected the daily
calls of the three LLMs with it since February 2026, and designed the rules of the game-theoretic combination
\gtTwo{} and the role-split policy \meTwo. The author made the
manual calls, posed the research question, chose the claims and the tests, and approved each result before it
entered the paper. This study builds on the author's earlier work on multi-LLM investment signals and game-theoretic
readings of their disagreement \cite{zhang2026benchmarking,zhang2026gametheoretic}.

\emph{Use of AI tools.} In line with IEEE policy, the author discloses that AI tools assisted with the writing of
this paper. Claude (Anthropic) was used to write analysis code and to draft and edit text under the author's
direction, and DeepSeek was used as an automated reviewer. All analysis code is scripted and reruns from the
ledger; the author checked the results and the text and takes full responsibility for the content.
\bibliographystyle{IEEEtran}
\bibliography{references}
\end{document}